\documentclass{article} 
\usepackage{iclr2027_conference,times}

\usepackage{amsmath,amsfonts,bm}

\def\eqref#1{equation~\ref{#1}}

\def\1{\bm{1}}

\DeclareMathAlphabet{\mathsfit}{\encodingdefault}{\sfdefault}{m}{sl}
\SetMathAlphabet{\mathsfit}{bold}{\encodingdefault}{\sfdefault}{bx}{n}

\usepackage{hyperref}
\usepackage{url}

\usepackage{bbm}
\usepackage{enumitem}
\usepackage{amssymb}
\usepackage{booktabs}
\usepackage{tabularx}
\usepackage[table]{xcolor}
\usepackage{array}
\usepackage{caption}
\usepackage{tcolorbox}
\usepackage{amsthm}
\usepackage{bm}
\usepackage{wrapfig}
\usepackage[utf8]{inputenc}
\usepackage{hyperref}
\usepackage{pifont}
\usepackage{graphicx}
\usepackage{xcolor}
\usepackage{booktabs}
\usepackage{rotating}
\usepackage{multirow}
\usepackage{booktabs}
\usepackage{multirow}
\usepackage[table]{xcolor}
\usepackage{graphicx}
\usepackage{fontawesome5}
\usepackage{minitoc}
\usepackage{titletoc}
\usepackage{float}

\usepackage[hyphenation]{impnattypo}

\DeclareRobustCommand{\mcicon}{\faIcon{list-ul}}
\DeclareRobustCommand{\oeicon}{\faIcon{pen}}

\newcommand{\cmark}{%
  \raisebox{-0.04em}{\textcolor{green!60!black}{\large\faCheckCircle}}%
}
\newcommand{\xmark}{%
  \raisebox{-0.2em}{\textcolor{black}{\large\faTimes}}%
}

\definecolor{projblue}{HTML}{2563EB}
\definecolor{softblack}{HTML}{333333}

\newcommand{\DATANAME}{\textsc{STRIDE}}

\title{\DATANAME{}: Evaluating Spatiotemporal Reasoning in Driving Edge Cases}

\author{%
\begin{minipage}[t]{\dimexpr\textwidth-2\tabcolsep\relax}
\centering
\vspace*{6pt}%
Lieqi Liu$^1$\thanks{\hspace{0.5mm}Equal contribution.},
Rui Gao$^1$\footnotemark[1],
Jia-Chen Gu$^1$,
Wenbo Hu$^1$
\\
Zhaobin Mo$^2$,
Ahmadreza Moradipari$^2$,
Nejib Ammar$^2$,
Wei Wang$^1$,
Nanyun Peng$^1$
\\[3pt]
{\normalfont
$^1$University of California, Los Angeles
\qquad
$^2$Toyota InfoTech Labs
}
\\
{\normalfont
\texttt{\{lieqiliu, rgao727, gujc, wenbohu\}@ucla.edu}
}
\\[3pt]
{\normalfont\normalsize
\href{https://pluslabnlp.github.io/STRIDE/}{
  \textcolor{projblue}{\faGlobe}\hspace{2pt}Website
}
\quad
\href{https://github.com/PlusLabNLP/STRIDE}{
  \textcolor{softblack}{\faGithub}\hspace{2pt}Code
}
}
\end{minipage}%
}

\iclrfinalcopy 
\begin{document}

\maketitle

\begin{abstract}

In autonomous driving, accurately perceiving surrounding objects is far from enough: failing to understand how they move and interact over time can lead to dangerous decisions. Evaluating vision-language models (VLMs) for autonomous driving therefore requires assessing their ability to connect spatial observations of past events with future consequences. Existing benchmarks usually oversimplify the problems by assessing driving edge cases only through basic perceptual questions without spatiotemporal considerations, or evaluating temporal understanding under simple scenarios, leaving the spatiotemporal reasoning required in complex driving scenarios insufficiently tested. 
To address this gap, we introduce \DATANAME{} (\textbf{S}patio\textbf{T}emporal \textbf{R}easoning \textbf{I}n \textbf{D}riving \textbf{E}dge Cases), an object-centric benchmark for evaluating spatiotemporal reasoning in long-tail and safety-critical scenarios. \DATANAME{} defines six task families: spatial perception, spatial understanding, temporal memory, temporal extrapolation, trajectory prediction, and scene-context awareness. 
It pairs multiple-choice questions covering a broad range of spatiotemporal judgments with related free-form visual question answering (VQA) that examines the same capabilities through detailed scene explanations and justifications of driving decisions. We curate \DATANAME{} by prioritizing scenes with unusual object behaviors and complex traffic interactions, constructing reference answers from structured dataset annotations through automated procedures, and applying limited language model assistance alongside human verification to guarantee annotation quality. Evaluation of 12 proprietary, open-weight, and driving-specialized models reveals substantial gaps: even GPT-6-Astra only achieves the highest average multiple-choice accuracy of 43.4\%, compared with a random-choice baseline of approximately 20\%. 
Our analysis further reveals that strong performance in describing driving scenes does not necessarily translate to accurate safety-critical judgments, and that models struggle more with reconstructing object motion than recalling past positions. These insights highlight \DATANAME{}'s benchmarking and diagnostic value for uncovering spatiotemporal driving limitations that aggregate VQA performance can conceal.

\end{abstract}
\section{Introduction}

\begin{figure*}[t]
  \centering

\includegraphics[width=\textwidth]{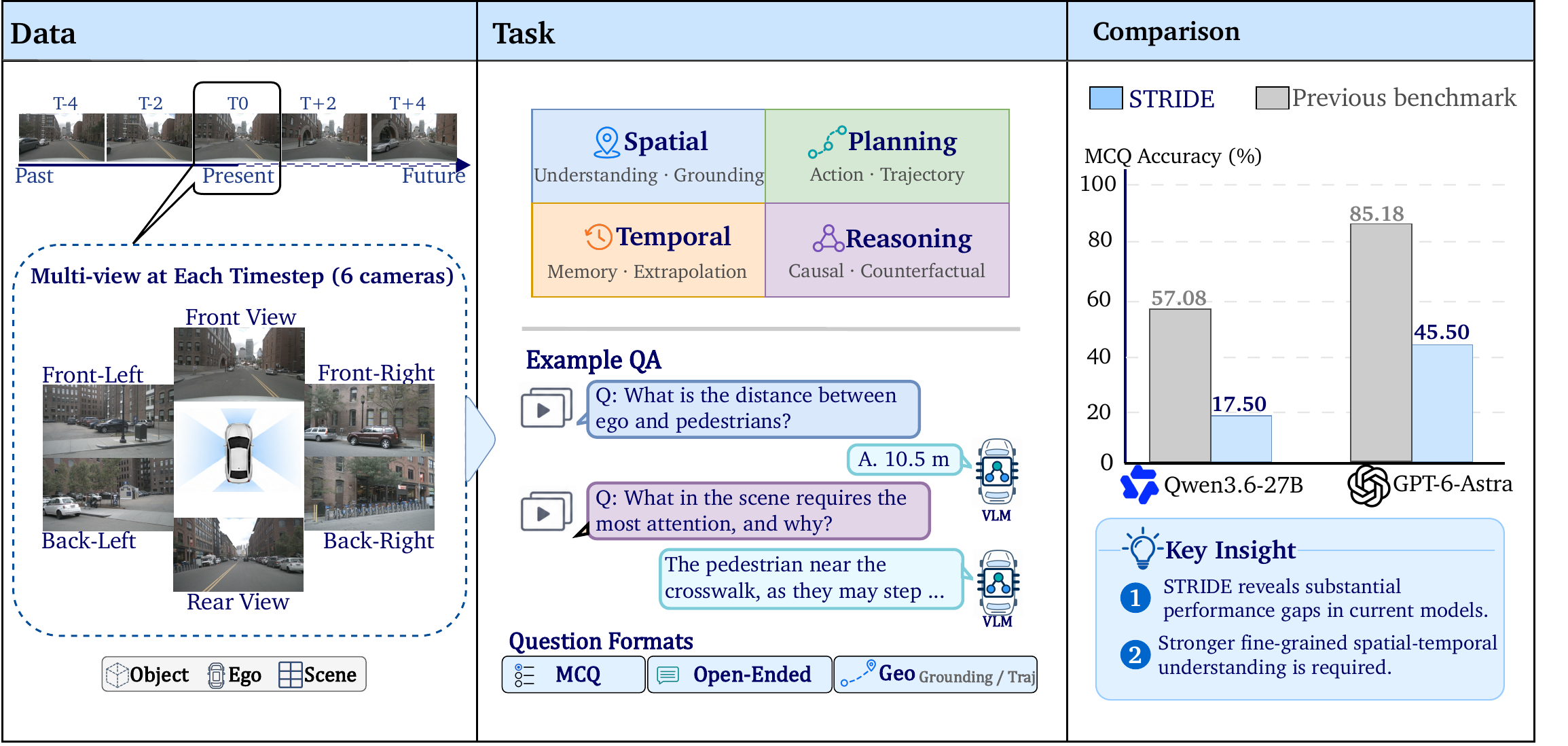}
  \caption{Overview of \DATANAME{}. Built from multi-view temporal driving data, \DATANAME{} evaluates spatiotemporal capabilities at the object and scene levels through complementary tasks. Qwen3.6-27B and GPT-6-Astra achieve substantially lower multiple-choice questions (MCQ) accuracy on \DATANAME{} than on the previous benchmark \citep{fruhwirth2026stsbench}.}
  \label{fig:flagship}
  \vspace{-2em}
\end{figure*}

Vision-language models (VLMs) demonstrate remarkable reasoning capabilities \citep{hwang2025emma,chi2026impromptu}, yet they still face significant challenges in the core demands of autonomous driving -- spatiotemporal perception, behavior prediction, and safe decision-making \citep{schwarting2018planning,li2024ego,chen2025automated,xie2025vlms,zhao2026extending}.
These limitations are particularly consequential in safety-critical scenarios that are underrepresented in driving data dominated by routine situations \citep{xu2026wod}. 
Resolving these complex, long-tail hazards requires continuous spatiotemporal reasoning: the ability to integrate past observations with shifting spatial dynamics to anticipate emerging risks. Because current models are largely untested against these specific demands, there is a critical need to systematically evaluate VLM spatiotemporal reasoning in high-stakes environments for autonomous driving.

Existing benchmarks for autonomous driving do not adequately assess whether VLMs can integrate spatial observations with past events to anticipate future consequences in challenging driving scenarios. 
Edge-case benchmarks often emphasize recognition questions that require limited reasoning about motion history or future interactions \citep{chen2025automated,chi2026impromptu}, while temporal reasoning benchmarks typically focus on comparatively simple scenarios, such as stationary vehicles and constant-speed straight driving \citep{cannons2025segments,fruhwirth2026stsbench}. 
As a result, VLMs’ ability to reason jointly about spatial and temporal relationships in scenarios involving unusual behaviors and complex interactions remains insufficiently tested. These gaps motivate a unified benchmark that evaluates spatiotemporal reasoning in such scenarios. 

To address this gap, we introduce \DATANAME{} (Figure~\ref{fig:flagship}), an object-centric benchmark for evaluating spatiotemporal reasoning in long-tail and safety-critical driving scenarios. 
We prioritize scenes with unusual object behaviors and complex traffic interactions, requiring models to locate nearby objects, reconstruct their past motion, and anticipate how their future behavior may affect the ego vehicle. 
To support fine-grained diagnosis of spatiotemporal reasoning, the benchmark comprises 39 manually designed question templates organized into six complementary task families: spatial perception, spatial understanding, temporal memory, temporal extrapolation, trajectory prediction, and scene-context awareness. 
Each evaluation combines object-centric temporal context with multiple-choice questions (MCQs) that test intermediate judgments and open-ended questions (OEQs) that assess their integration into driving explanations and decisions. 

Our evaluation of 12 proprietary, open-weight, and driving-specialized models highlights the difficulty of \DATANAME{}. Qwen3.6-27B \citep{qwen3.6-27b} and GPT-6-Astra both achieve substantially lower MCQ accuracy than on the previous temporal reasoning benchmark \citep{fruhwirth2026stsbench} (Figure~\ref{fig:flagship}). Fine-grained analysis reveals difficulty connecting observations across frames into a coherent account of object dynamics: GPT-5.5 achieves 72\% accuracy for previous relative position but only 18\% for recent motion patterns. 
Our evaluation also reveals gaps that aggregate scores can conceal. High scenario explanation VQA scores coexist with weak judgments about drivable space and specific risks, suggesting that broad scene caption tasks may partly rely on generic driving knowledge. This concern echoes findings that models retain substantial performance even when visual inputs are replaced with black images \citep{xie2025vlms}. These findings highlight \DATANAME{}'s diagnostic value and motivate evaluating driving judgments, temporal integration, and geometric prediction alongside verbal explanations.


\section{Related Work}

\paragraph{Driving VQA Benchmarks}
Driving vision-language benchmarks evaluate scene understanding through questions about road entities, vehicle behavior, and driving context \citep{caesar2020nuscenes,marcu2024lingoqa}. Beyond individual questions, subsequent efforts connect perception with prediction and planning through graph-structured QA, annotated reasoning chains, and counterfactual supervision \citep{sima2024drivelm,nie2024reason2drive,wang2025omnidrive}. Spatially grounded evaluation has also expanded toward multi-view understanding and explicit 3D localization \citep{wang2025omnidrive,zhao2026extending}. However, reliability studies show that favorable QA or planning scores can reflect textual shortcuts or reliance on ego-state information, motivating closer examination of the capabilities underlying model performance \citep{li2024ego,xie2025vlms}. 

\paragraph{Spatiotemporal and Long-Tail Evaluation}
Recent benchmarks increasingly assess temporal understanding through multi-frame traffic interactions, action recognition, event ordering, and temporal localization \citep{cannons2025segments,fruhwirth2026stsbench}. In parallel, corner-case and long-tail datasets broaden evaluation beyond routine driving, covering unusual objects, unstructured roads, and challenging agent interactions through perception, driving suggestions, and trajectory tasks \citep{chen2025automated,chi2026impromptu,xu2026wod}. These efforts establish complementary foundations for evaluating temporal capabilities and scenario difficulty. 
\section{\DATANAME{}}
\label{sec:data}
\providecommand{\cmark}{\textcolor{green!55!black}{\ding{51}}}
\providecommand{\pmark}{\textcolor{orange!85!black}{\ding{108}}}
\providecommand{\xmark}{\textcolor{gray!65}{--}}
The \DATANAME{} benchmark is designed to evaluate the spatiotemporal reasoning of autonomous-driving agents in long-tail scenarios with reliable ground truth and interpretable assessment. 
As summarized in Table~\ref{tab:benchmark_comparison}, \DATANAME{} combines spatial and temporal reasoning, edge-case evaluation, explicit grounding, and ego-trajectory prediction with both multiple-choice and open-ended QA.

\begin{table*}[t]
\centering

\caption{
Comparison of representative autonomous-driving vision-language benchmarks.
\protect\cmark: explicitly evaluated; \protect\xmark: no dedicated evaluation reported.
 ``Mixed'' indicates dataset-dependent camera configurations.
\mcicon: multiple-choice or fixed-vocabulary short-answer QA;
\oeicon: open-ended QA.
A capability is marked only when explicitly evaluated: Spatial concerns where
objects are and how they relate, Temporal concerns reasoning across frames or
about what happens next, Edge Cases concerns unusual or safety-critical
situations, Grounding requires explicit localization, and Ego Traj. requires
future ego-trajectory coordinates.
}
\label{tab:benchmark_comparison}

\setlength{\tabcolsep}{3.2pt}
\renewcommand{\arraystretch}{1.18}

\resizebox{\textwidth}{!}{%
\begin{tabular}{lcccccccc}
\toprule

\multirow{2}{*}{\textbf{Benchmark}} &
\multicolumn{2}{c}{\textbf{Input}} &
\multirow{2}{*}{\textbf{QA Format}} &
\multicolumn{5}{c}{\textbf{Capabilities}} \\

\cmidrule(lr){2-3}
\cmidrule(lr){5-9}

&
\textbf{Format} &
\textbf{View} &
&
\textbf{Spatial} &
\textbf{Temporal} &
\textbf{Edge Cases} &
\textbf{Grounding} &
\textbf{Ego Traj.} \\

\midrule

NuScenes-QA \citep{qian2024nuscenes}
& Image / Video
& Surround
& \mcicon
& \cmark & \xmark & \xmark & \xmark & \xmark \\

DriveLM-nuScenes \citep{sima2024drivelm}
& Image
& Surround
& \mcicon\, \oeicon
& \cmark & \cmark & \xmark & \cmark & \cmark \\

Reason2Drive \citep{nie2024reason2drive}
& Video
& Mixed
& \oeicon
& \cmark & \cmark & \xmark & \cmark & \cmark \\

LingoQA \citep{marcu2024lingoqa}
& Video
& Front
& \oeicon
& \cmark & \cmark & \xmark & \xmark & \xmark \\

OmniDrive \citep{wang2025omnidrive}
& Image
& Surround
& \oeicon
& \cmark & \cmark & \cmark & \xmark & \cmark \\

DriveBench \citep{xie2025vlms}
& Image
& Surround
& \mcicon\, \oeicon
& \xmark & \cmark & \cmark & \cmark & \xmark \\

NuPlanQA-Eval \citep{park2025nuplanqa}
& Image
& Surround
& \mcicon
& \cmark & \xmark & \xmark & \xmark & \xmark \\

STSBench \citep{fruhwirth2026stsbench}
& Video
& Surround
& \mcicon
& \cmark & \cmark & \xmark & \xmark & \xmark \\

\midrule

\rowcolor{green!8}
\textbf{STRIDE}
& \textbf{Video}
& \textbf{Mixed}
& \textbf{\mcicon\, \oeicon}
& \cmark & \cmark & \cmark & \cmark & \cmark \\

\bottomrule
\end{tabular}%
}

\vspace{3pt}

\begin{minipage}{\textwidth}
\footnotesize
\setlength{\parindent}{0pt}
\setlength{\parskip}{2pt}

\end{minipage}

\end{table*}

\subsection{Task Definition}
\label{sec:Task Definition}


Autonomous driving is an anticipatory decision-making problem: an agent must understand the previous and current scene, predict how it may evolve, and take preventive action before developing risks lead to dangerous interactions. Such anticipation of recognizing emerging conflicts
and determining when and how the ego vehicle should respond depends jointly on spatial and temporal reasoning. Object positions, distances, and available driving space establish the current conditions for action, while changes across consecutive frames reveal motion trends and support predictions of future interactions. Motivated by these requirements, we organize \DATANAME{} around the complementary dimensions of \emph{space} and \emph{time}.

\vspace{-2mm}
\paragraph{Space} This dimension is divided into two hierarchical levels of increasing difficulty, from individual object location to multi-object relation. \textbf{Spatial perception} evaluates basic object-level properties that are directly observable from the scene, including relative position, distance, direction of motion, and lane occupancy. \textbf{Spatial understanding} builds on these observations and evaluates whether models can interpret the driving significance of these spatial relationships. Tasks include determining whether an object restricts the drivable region, affects lane-change feasibility, constrains the ego vehicle's motion, or poses a potential risk. 

\vspace{-2mm}
\paragraph{Time} This dimension extends this representation of the current scene backward and forward in time. Taking the current spatial configuration as the reference state, \textbf{temporal memory} evaluates whether a model can reconstruct how that state developed from previous observations. This includes retrieving previously visible or disappeared objects, earlier object locations, occlusion and reappearance, and recent motion trends. In the opposite direction, \textbf{temporal extrapolation} evaluates whether the model can anticipate how the current scene may evolve, including future object motion, lane occupancy, collision timing, and likely subsequent events. Together, these two task families test whether a model can connect the present scene to its recent history and possible future.

\vspace{-2mm}
\paragraph{Driving Context}
Above tasks primarily assess capabilities through questions grounded in a target object. To account for the ego vehicle's own motion and the broader environment in which object interactions occur, we therefore include the evaluation on \textbf{trajectory prediction} and \textbf{scene-context awareness}. Trajectory tasks assess the ego vehicle's past and future motion. Scene-context tasks assess environmental conditions, road structure, and surrounding activity that provide context for interpreting individual objects and their interactions. 

\subsection{Data Construction}

\begin{figure*}[t]
  \centering
  \includegraphics[width=\textwidth]{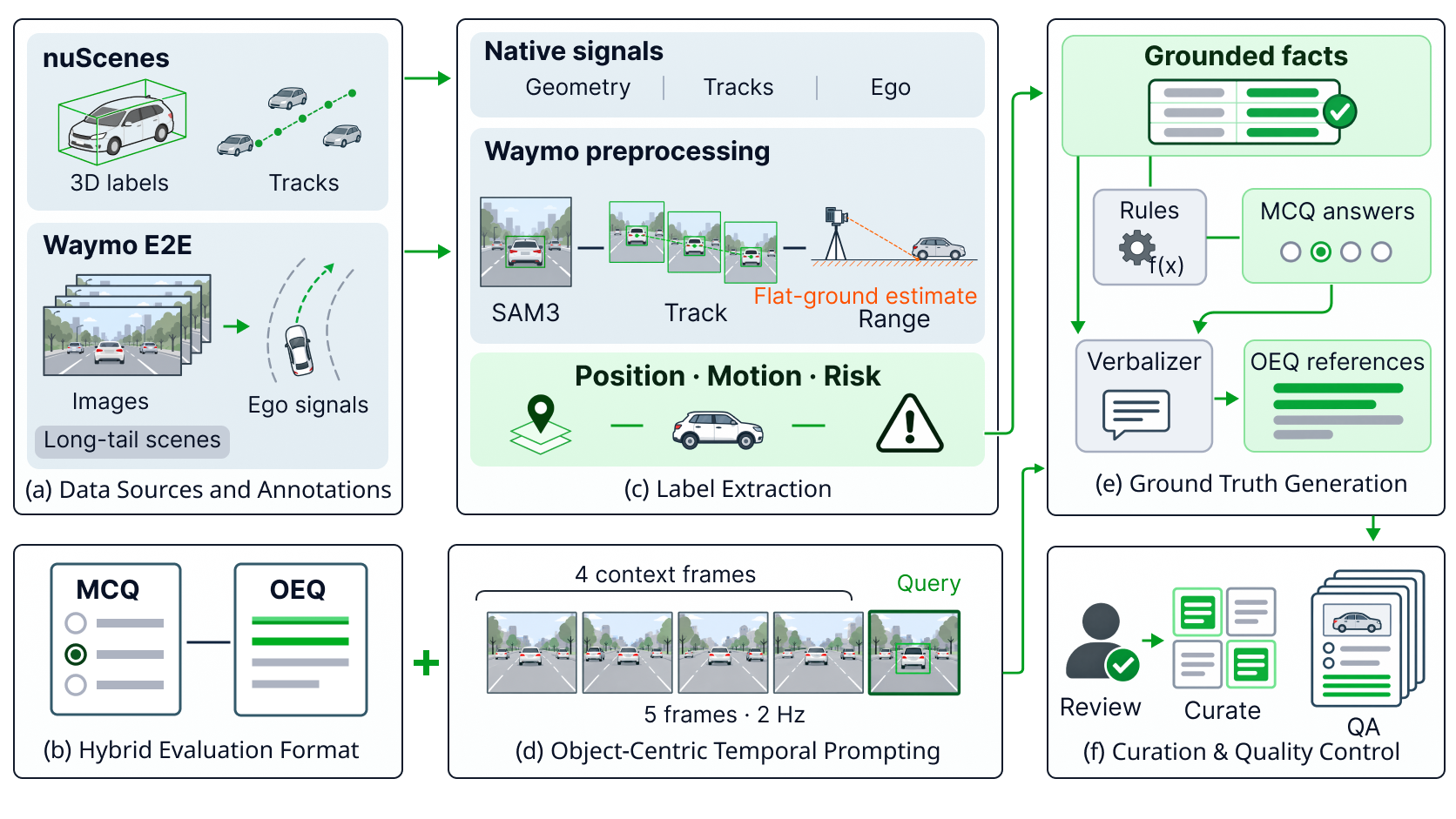}
  \caption{Data generation pipeline of \DATANAME{}. From left to right, dataset annotations and extracted signals are converted into structured labels. Manually designed hybrid format pair these labels with five-frame context to generate linked MCQ answers and verbalized OEQ references, followed by human review and curation.}
  \label{fig:pipeline}
\end{figure*}

\paragraph{Data Source and Annotation}
We construct \DATANAME{} from two complementary sources: \textbf{nuScenes} \citep{caesar2020nuscenes} and \textbf{WOD-E2E} \citep{xu2026wod}. As seen in Figure ~\ref{fig:pipeline}(a), each of them provides different annotations. nuScenes is a 
multimodal autonomous-driving dataset comprising 1,000 driving scenes of 20 seconds each, with six-camera surround-view coverage and 3D object annotations. 
While providing broad coverage of everyday urban driving and rich annotations, such naturally collected datasets are dominated by nominal driving situations \citep{zhai2023rethinking,li2024ego,dauner2024navsim,pmlr-v270-tian25b}. 
We complement nuScenes with WOD-E2E, a more recent dataset curated to emphasize challenging long-tail driving scenarios, including unusual agent interactions, cut-ins, construction zones, encounters with vulnerable road users, unexpected obstacles, and other safety-critical situations.
For WOD-E2E, we supplement the released annotations with object-centric signals derived through detection, tracking, and geometric distance estimation, as detailed in Appendix~\ref{sec:waymo-preprocessing}.

\vspace{-2mm}
\paragraph{Data Sampling Strategy}
To keep evaluation computationally manageable while preserving scenario diversity, we construct a compact evaluation set through template-based curation. For each question template, we sample up to 50 valid instances from scenes in which the required annotations and visual evidence are available. Since the two source datasets provide different annotation types and scenario characteristics, not every template is applicable to both datasets due to missing annotations required. To address this issue and ensure that our ground truth is trustworthy, each template is instantiated only on the dataset or datasets that provide sufficient information to support reliable ground-truth generation. Within the eligible candidates, we prioritize informative and challenging examples using a combination of complementary signals. These include scene-intrinsic factors such as traffic-participant density and interactions, weather and visibility conditions, hazardous elements, unusual agent behavior, and overall scene complexity such as involvement of vulnerable road users. 

\vspace{-2mm}
\paragraph{Hybrid Evaluation}
\DATANAME{} combines multiple-choice and free-form VQA to assess shared spatial and temporal capabilities through complementary response formats (Figure ~\ref{fig:pipeline}(b)). Within each task family, questions are aligned around shared scene attributes, with the two formats evaluated independently. For example, the \textbf{spatial understanding} family pairs multiple-choice questions on physical constraints, available drivable space, lane-change feasibility, and risk with the free-form question: \textit{Describe the overall spatial situation of the ego vehicle in the scene, including existing physical constraints, available drivable space, risk level, and feasible maneuvers.} The multiple-choice questions serve as intermediate judgments of individual attributes, which are then integrated in the linked open-ended question for a broader description or justification of the driving situation. This alignment supports complementary analysis of judgment accuracy and explanatory quality.

\begin{figure*}[t]
  \centering
  \includegraphics[width=\textwidth]{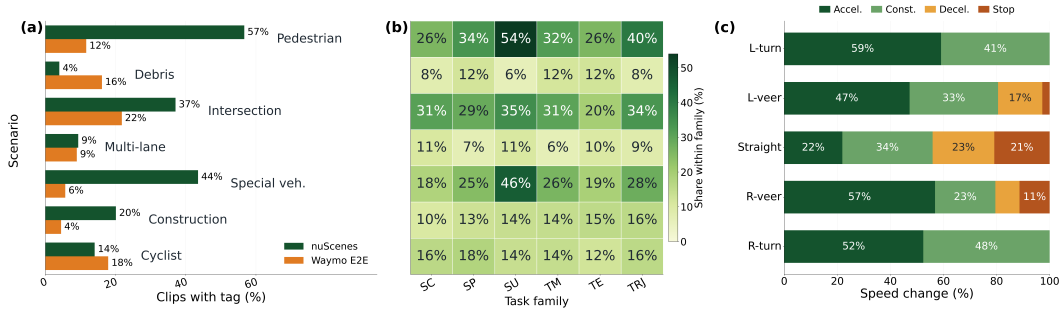}
  \caption{\textbf{Coverage of \DATANAME.}
  \textbf{(a)} Share of tagged clips containing each challenging situation in
  nuScenes and Waymo~E2E.
  \textbf{(b)} Share of clips within each task family carrying the same scenario
  tag (SC: scene context, SP: spatial perception, SU: spatial understanding,
  TM: temporal memory, TE: temporal extrapolation, TRJ: trajectory). Tags are
  multi-label, so columns need not sum to $100\%$.
  \textbf{(c)} Future ego-maneuver composition by speed change over unique
  ground-truth motion patterns; bars are normalized within each maneuver.}
  \label{fig:dataset-coverage}
\end{figure*}

\vspace{-2mm}
\paragraph{Quality Control} After construction, we remove question types lacking sufficient source annotations for reliable ground truth. OEQ references are further refined using human scores and VLM assisted error analysis to correct unsupported or incomplete answers. After human review of the selected subset, we replace weak references with stronger model-assisted answers based on human scoring of the reference answer and the corresponding model response generated by GPT-5.5. More details can be found in Appendix ~\ref{sec:quality_control}.

\subsection{Dataset Statistics}
Our benchmark comprises $2{,}350$ questions spanning six task families and $1{,}452$ distinct clips. Of these, $1{,}150$ questions are derived from nuScenes~\citep{caesar2020nuscenes} using $23$ templates, while $1{,}200$ come from Waymo~E2E~\citep{xu2026wod} using $24$ templates; $1{,}700$ are multiple-choice and $650$ are open-ended.

Figure~\ref{fig:dataset-coverage}(a) highlights the complementary corner-case coverage: nuScenes contains a larger proportion of pedestrian, intersection, special-vehicle, and construction scenarios, whereas Waymo~E2E contributes relatively more foreign-object and cyclist cases. Figure~\ref{fig:dataset-coverage}(b) shows that these tags are not confined to particular question types. Pedestrians, intersections, and special vehicles recur across all six families, so nearly every capability is evaluated under challenging conditions. Behaviorally, the benchmark includes various future-motion patterns (Figure~\ref{fig:dataset-coverage}(c)). Pairing every maneuver with multiple speed profiles prevents models from inferring speed from motion direction alone.

\section{Experiments}
\label{sec:experiments}

\subsection{Experimental Setup}

We evaluate \DATANAME{} with a diverse set of VLMs spanning proprietary general-purpose models including GPT-6-Astra, Gemini-3.1-Pro, and Claude-Sonnet-5, open-weight models of different scales including Qwen3-VL-8B-Thinking, Qwen3.6-27B, and Qwen3.6-35B-A3B, and driving-specialized models including Dolphins and UniAD.
This model suite allows us to compare both broad multimodal capabilities and domain-specific expertise under a unified evaluation protocol.

\vspace{-2mm}
\paragraph{Implementation Details}
All models are evaluated using the same benchmark inputs and prompting protocol whenever their interfaces permit. Open-weight models are downloaded from \texttt{HuggingFace} and evaluated using \texttt{vLLM} \citep{kwon2023efficient} on NVIDIA A6000 GPUs, while proprietary models are accessed through their official APIs. Unless otherwise stated, we use deterministic or near-deterministic decoding settings to reduce evaluation variance.
More details about experimental settings can be found in Appendix~\ref{sec:additional_experiments}.

\begin{table*}[t]
\centering
\caption{
Results on the nuScenes and Waymo E2E splits of \DATANAME{}.
MC accuracy is reported in percentage. OEQ performance is reported with scale 0-10. 
Best and second-best model results are shown in \textbf{bold} and \underline{underline}, respectively.
}
\label{tab:main_results}
\small
\setlength{\tabcolsep}{3.8pt}
\begin{tabular}{lcccc|cccc}
\toprule
& \multicolumn{4}{c|}{\textbf{nuScenes}} 
& \multicolumn{4}{c}{\textbf{Waymo E2E}} \\
\cmidrule(lr){2-5}
\cmidrule(lr){6-9}

\textbf{Model} &
\textbf{MCQ $\uparrow$} &
\textbf{OEQ $\uparrow$} &
\textbf{ADE $\downarrow$} &
\textbf{FDE $\downarrow$} &
\textbf{MCQ $\uparrow$} &
\textbf{OEQ $\uparrow$} &
\textbf{ADE $\downarrow$} &
\textbf{FDE $\downarrow$} \\

\midrule

\rowcolor{gray!15}
\textsc{Random Chance}
& 19.8 & -- & -- & --
& 20.3 & -- & -- & -- \\

\rowcolor{gray!15}
\textsc{Human Performance}
& 79 & -- & -- & --
& 69 & -- & -- & -- \\

\midrule

\rowcolor{gray!15}\multicolumn{9}{l}{\textsc{Driving Expert Models}} \\

Dolphins \citep{ma2024dolphins}
& 18.5 & 3.24 & 8.16 & 3.81
& 22.2 & 1.86 & 18.43 & \underline{8.60} \\

UniAD 2.0 \citep{hu2023planning}
& -- & -- & \textbf{0.56} & \textbf{0.79}
& -- & -- & -- & -- \\

\midrule

\rowcolor{gray!15}\multicolumn{9}{l}{\textsc{Open-weight Models}} \\

Qwen3.6-27B
& 17.5 & 5.99 & 5.81 & 4.06
& 28.4 & 5.87 & 20.37 & 19.25 \\

Qwen3-VL-8B-Thinking
& 23.4 & 5.50 & 8.00 & 3.80
& 32.0 & 4.72 & 22.36 & 19.37 \\

Qwen3-VL-30B-A3B-Thinking
& 26.8 & 5.45 & 7.12 & 3.42
& 31.8 & 4.98 & \textbf{11.83} & \textbf{6.18} \\

Qwen3.6-35B-A3B
& 31.6 & 5.32 & 7.42 & 5.45
& 34.3 & 5.83 & 18.55 & 8.86 \\

\midrule

\rowcolor{gray!15}\multicolumn{9}{l}{\textsc{Proprietary Models}} \\

Claude-Sonnet-5
& 29.4 & 6.42 & 0.87 & 1.30
& 39.2 & \textbf{7.01} & \underline{13.16} & 10.27 \\

GPT-5.6-Sol
& 38.5 & 5.65 & 0.69 & 0.99
& 37.7 & 6.48 & 13.81 & 12.39 \\

Gemini-3.1-Pro
& 37.8 & 4.53 & 1.42 & 1.84
& \underline{39.7} & 4.03 & 14.22 & 14.21 \\

Gemini-3.6-Flash
& 38.9 & \underline{6.50} & 0.69 & 1.01
& 38.8 & 6.20 & 13.53 & 11.37 \\

GPT-5.5
& \underline{39.3} & \textbf{6.68} & 0.65 & \underline{0.97}
& 38.7 & \underline{7.00} & 13.84 & 12.57 \\

GPT-6-Astra
& \textbf{45.5} & 6.41 & \underline{0.60} & 0.98
& \textbf{41.2} & 6.77 & 13.64 & 12.03 \\

\bottomrule
\end{tabular}
\end{table*}

\vspace{-2mm}
\paragraph{Evaluation Metrics.}
For multiple-choice tasks, we report accuracy. For open-ended questions, we report the mean score assigned by a vision-language judge on a 0--10 scale.
For trajectory prediction, each predicted waypoint is compared with the ground-truth waypoint at the same timestep by Euclidean ($\ell_2$) distance.
We report two summaries of that distance: Average Displacement Error (ADE), the mean $\ell_2$ over timesteps, and Final Displacement Error (FDE), the $\ell_2$ at the final timestep. More details about scoring can be found in Appendix ~\ref{sec:app_scoring}

\subsection{Main Results}

Table~\ref{tab:main_results} reveals substantial limitations even in the strongest evaluated models.
GPT-6-Astra achieves the highest MCQ accuracy on both nuScenes and Waymo E2E, yet reaches only 45.5\% and 41.2\%, respectively. 
Although above the random-choice baselines of 19.8\% and 20.3\%, these results leave more than half of the questions incorrectly answered, highlighting persistent difficulties with the benchmark's spatial and temporal judgments. 
For OEQ evaluation, performance also varies across response formats and tasks. GPT-5.5 leads on nuScenes with a score of 6.68, while Claude-Sonnet-5 achieves the highest score on Waymo E2E at 7.01, closely followed by GPT-5.5 at 7.00.
Meanwhile, the driving expert UniAD 2.0 excels at trajectory prediction on nuScenes, achieving an ADE/FDE of 0.56/0.79, compared with 0.60/0.98 for GPT-6-Astra and 0.65/0.97 for GPT-5.5. 
In summary, these results expose a key limitation: strong aggregate VQA performance does not establish reliable spatiotemporal judgment or superior trajectory prediction. Improving the consistency of these capabilities remains an important direction for future work.
\section{Analysis}
\label{sec:analysis}

We examine the models in Table~\ref{tab:main_results} across four capability categories using the selected per-task results in Table~\ref{tab:analysis_summary}. Complete results can be found in Appendix Tables~\ref{tab:nusc_subtask_results} and~\ref{tab:waymo_subtask_results}. Scene and risk interpretation are discussed under Section~\ref{sec:spatial_understanding}, while future-trajectory tasks are included under Section~\ref{sec:extrapolation}.

\begin{table}[t]
\centering
\caption{
Representative per-task results used in the analysis.
GPT-5.6: GPT-5.6-Sol; G-Fl.: Gemini-3.6-Flash; G-Pro: Gemini-3.1-Pro;
Claude: Claude-Sonnet-5; Q35: Qwen3.6-35B-A3B;
Q30-T/Q8-T: Qwen3-VL-30B-A3B-Thinking/Qwen3-VL-8B-Thinking.
}
\label{tab:analysis_summary}

\small
\setlength{\tabcolsep}{4.2pt}
\renewcommand{\arraystretch}{1.08}

\begin{tabular*}{\linewidth}{
@{\extracolsep{\fill}}
llcccc
}
\toprule
\textbf{(a) nuScenes}
& \textbf{Model}
& \multicolumn{4}{c}{\textbf{Representative tasks}} \\
\cmidrule(lr){3-6}
\textbf{Capability}
&
& \textbf{Task 1}
& \textbf{Score}
& \textbf{Task 2}
& \textbf{Score} \\
\midrule

Spatial perception
& G-Pro
& SP-3: Distance Estimate
& 64\%
& SP-7: Spatial desc.
& 2.10 \\

Spatial understanding
& GPT-5.5
& SC-6: Scene desc.
& 9.22
& SU-2: Drivable region
& 22\% \\

Temporal memory
& GPT-5.5
& TM-2: Prev. position
& 72\%
& TM-3: Recent motion
& 18\% \\

Temporal extrapolation
& Best
& TE-1: Future motion
& 40\%
& TE-6: Scene evolution
& 6.98 \\

\bottomrule
\end{tabular*}

\vspace{4pt}

\begin{tabular*}{\linewidth}{
@{\extracolsep{\fill}}
llcccc
}
\toprule
\textbf{(b) WOD-E2E}
& \textbf{Model}
& \multicolumn{4}{c}{\textbf{Representative tasks}} \\
\cmidrule(lr){3-6}
\textbf{Capability}
&
& \textbf{Task 1}
& \textbf{Score}
& \textbf{Task 2}
& \textbf{Score} \\
\midrule

Scene/risk reasoning
& GPT-5.5
& SC-4: Scene/risk/action
& 8.08
& SC-2: Risk source
& 28\% \\

Spatial perception
& G-Fl. / Claude
& SP-3: Distance Estimate
& 62\%
& SP-3a: Metric distance
& 6.28 \\

Temporal memory
& Q30-T
& TM-2: Prev. position
& 10\%
& TM-3: Recent motion
& 32\% \\

Temporal extrapolation
& Best
& TE-1: Future motion
& 36\%
& TE-5: Next event
& 34\% \\

\bottomrule
\end{tabular*}

\vspace{4pt}

\begin{tabular*}{\linewidth}{
@{\extracolsep{\fill}}
llcccc
}
\toprule
\textbf{(c) Trajectory}
& \textbf{Split / Task}
& \textbf{Parsing Rate}
& \textbf{ADE valid}
& \textbf{ADE all}
& \textbf{FDE all} \\
\midrule

Q35
& nuScenes / TRJ-6
& 16\%
& 0.32
& 8.19
& 4.61 \\

Dolphins
& WOD-E2E / TRJ-5,6
& 0\%
& --
& 18.43
& 8.60 \\

GPT-5.5
& WOD-E2E / TRJ-5
& 100\%
& 23.44
& 23.44
& 20.10 \\

\bottomrule
\end{tabular*}

\end{table}

\subsection{Spatial Perception}

Performance varies across spatial attributes and response formats. On nuScenes, the best relative-position and distance accuracies are 62\% and 64\%, compared with 38\% for trajectory direction and 44\% for occlusion state. Integrated spatial descriptions remain difficult: GPT-5.5 leads with 3.66/10, while Gemini-3.1-Pro scores 2.10/10 despite leading distance MCQs. On Waymo E2E, distance MCQ and OEQ leaders also differ: Gemini-3.6-Flash achieves 62\% accuracy, whereas Claude-Sonnet-5 scores 6.28/10. Success on individual spatial attributes therefore does not consistently extend across response formats or to descriptions integrating multiple relations.

Output-format compliance further affects measured performance. For Qwen3.6-27B on nuScenes SP-3, only 34\% of responses are of the valid format; accuracy is 22\% overall and 64.7\% among the 17 parsed responses. This suggests insufficient instruction-following, although performance on this selected subset cannot establish accuracy on the full task.

\subsection{Spatial Understanding}
\label{sec:spatial_understanding}


Our benchmark further reveals that strong scene descriptions do not ensure reliable driving judgments. On nuScenes, GPT-5.5 scores 9.22/10 for scene description but only 22\% for both drivable-region identification and lane-change clearance. Qwen3-VL-8B-Thinking also exceeds all proprietary models on physical constraints, reaching 42\%.

On Waymo E2E, GPT-5.5 scores 8.08/10 for scene/risk/action explanations, alongside accuracies of 44\%, 28\%, and 48\% for scene classification, risk identification, and preferred action. Fine-grained prompts and task-specific scoring criteria in \DATANAME{} make these comparisons more informative by separating broad explanatory quality from precise driving judgments. As a result, differences between OEQ and MCQ performance should not be interpreted as contradictory answers to equivalent questions, but rather as complementary evidence about distinct aspects of model capability.

\subsection{Temporal Memory}


Recalling earlier positions is easier than reconstructing motion on nuScenes. GPT-5.5 and GPT-5.6-Sol both achieve 72\% for previous object position but only 18\% and 16\%, respectively, for recent motion, near its 16.7\% chance baseline. Memory of currently occluded objects is also limited: Qwen3.6-35B-A3B leads at 36\%, against a 25\% baseline. These results suggest difficulty integrating observations into a consistent motion history.

Waymo E2E shows a less uniform pattern: Qwen3-VL-30B-A3B-Thinking achieves 32\% for recent motion but only 10\% for previous position. Past ego-maneuver classification is comparatively stronger, reaching 74\% for Claude-Sonnet-5. Strong ego-motion summaries therefore do not necessarily translate into reliable object-level memory, particularly when models must maintain object identities and reconstruct motion histories across multiple frames.

\begin{figure*}[t]
    \centering
    \includegraphics[width=\textwidth]{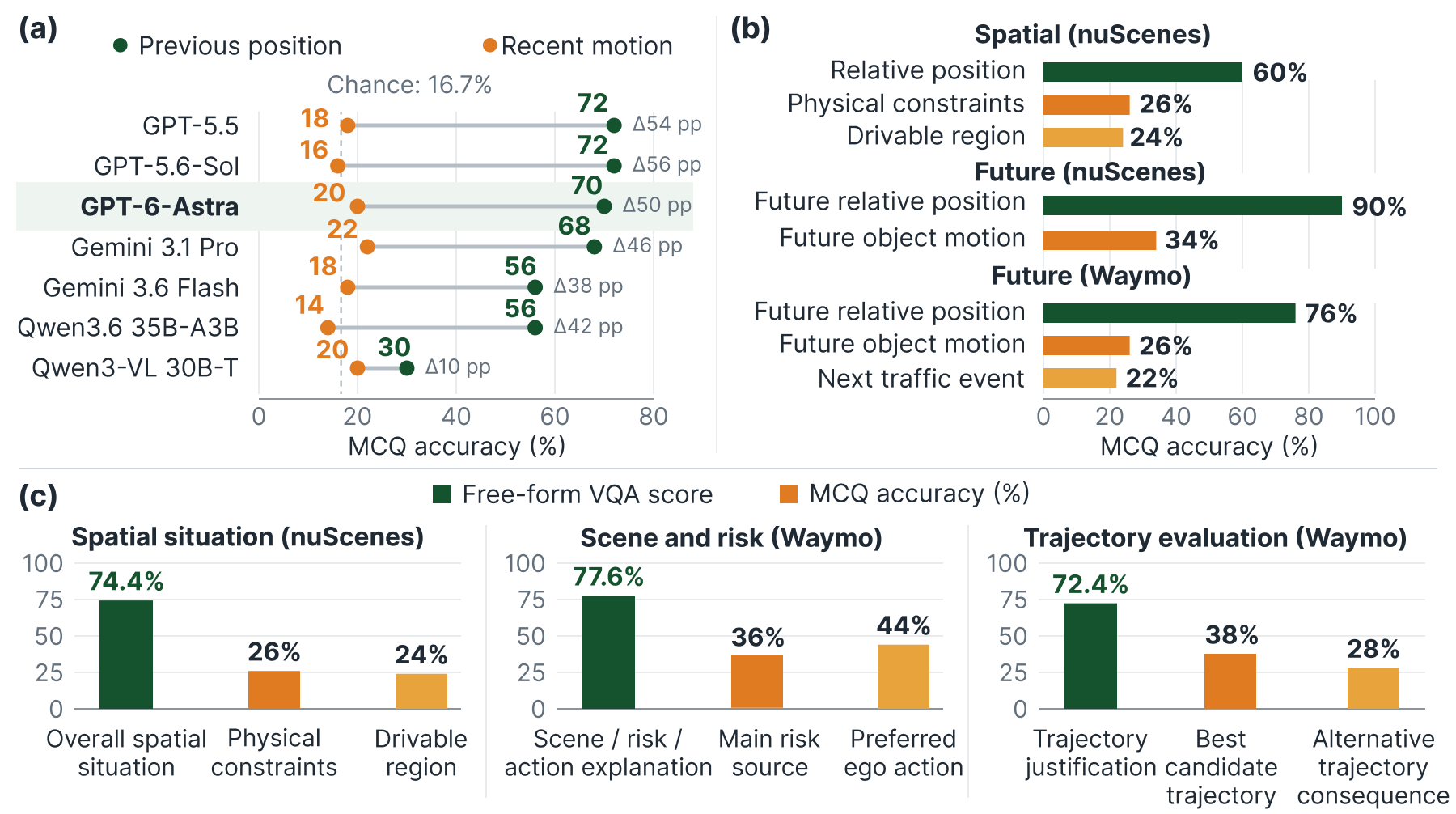}
    \caption{
    Analysis of GPT-6-Astra's driving capabilities.
    (a) Accuracy gaps between recalling previous object positions and reconstructing recent motion; $\Delta$ denotes the difference in percentage points.
    (b) Stronger positional judgments along with weaker spatial-constraint and future-event reasoning.
    (c) Favorable explanation scores with low accuracy on related driving judgments.
    Judge scores are multiplied by 10 for visualization on a shared 0--100 scale.
    }
    \label{fig:gpt6_astra_failure_modes}
\end{figure*}

\subsection{Temporal Extrapolation}
\label{sec:extrapolation}

Anticipating behavior and its driving consequences remains difficult. On nuScenes, the best future-object-motion accuracy is 40\%, and the best scene-evolution OEQ score is 6.98/10. On Waymo E2E, the best accuracies for future object motion, ego response, and the next traffic event are 36\%, 34\%, and 34\%. Trajectory evaluation also yields different rankings: Qwen3-VL-30B-A3B-Thinking achieves the lowest Waymo ADE/FDE of 11.83/6.18 despite weaker aggregate VQA performance.

Parsing success is essential to interpreting displacement errors. On nuScenes TRJ-6, Qwen3.6-35B-A3B parses successfully on only 16\% of examples; ADE is 0.32 on valid outputs but 8.19 overall. Dolphins parses 18\% of outputs on each nuScenes waypoint task and none on Waymo. Its finite Waymo FDE of 8.60 reflects the stationary-at-origin fallback for invalid outputs, rather than successful waypoint generation. Conversely, GPT-5.5 achieves 100\% parsing on Waymo TRJ-5 but still incurs an ADE of 23.44. Format and trajectory accuracy therefore require joint assessment.

\subsection{A Look into Failure Cases of GPT-6-Astra}

Although GPT-6-Astra achieves the highest aggregate MCQ accuracy on both nuScenes and Waymo E2E split of \DATANAME, its fine-grained results reveal several remaining limitations, most of which apply to all the VLMs we have tested. 

\vspace{-2mm}
\paragraph{Weak spatial reasoning and temporal tracking across frames}
GPT-6-Astra performs substantially better at identifying an object's state at a particular moment than at reconstructing how that state changes over time. As shown in Figure~\ref{fig:gpt6_astra_failure_modes}(a), it achieves 70\% accuracy for previous object position (TM-2) on nuScenes, but only 20\% for recent motion pattern (TM-3), a gap of 50 percentage points. This pattern extends across all seven models shown, with gaps ranging from 10 to 56 percentage points, suggesting a shared limitation in integrating observations across frames. Despite differences in positional recall, recent-motion accuracy remains consistently low at 14--22\%, close to the 16.7\% chance baseline. Future research could investigate memory mechanisms that explicitly maintain object identities, update their states, and retrieve relevant motion history, enabling driving agents to use past observations more reliably as a step toward practical deployment.

Figure~\ref{fig:gpt6_astra_failure_modes}(b) contrasts GPT-6-Astra's positional judgments with its performance on spatial constraints and future developments. On nuScenes, relative-position accuracy reaches 60\%, compared with 26\% for physical constraints and 24\% for drivable regions. Models perform well on simple recognition questions but struggle with safety-critical, action-relevant judgments that require a deeper understanding of how the scene evolves over time.
A similar gap appears in future reasoning: future relative-position accuracy reaches 90\% on nuScenes and 76\% on Waymo E2E, whereas future object-motion accuracy is only 34\% and 26\%, respectively. On Waymo E2E, next-traffic-event accuracy further falls to 22\%. 

\vspace{-2mm}
\paragraph{Strong scene-description scores coexist with weak safety-critical judgments}
GPT-6-Astra receives favorable scores for integrated scene explanations while remaining unreliable on specific safety-critical questions, as illustrated in Figure~\ref{fig:gpt6_astra_failure_modes}(c). On nuScenes, its overall spatial-situation explanations receive an average judge score of 7.44/10, alongside the low physical-constraint and drivable-region accuracies noted above. On Waymo E2E, scene/risk/action explanations receive 7.76/10, yet MCQ accuracy is only 36\% for identifying the main risk source and 44\% for selecting the preferred ego action. This pattern also extends to trajectory evaluation: trajectory justifications receive 7.24/10, while best-candidate trajectory selection and alternative-trajectory consequence judgments achieve only 38\% and 28\% accuracy, respectively. These aggregate results suggest that evaluations centered on broad scene descriptions may present an overly optimistic picture of driving competence, obscuring weaknesses in the detailed capabilities required for safety-critical judgments.

\section{Conclusion}

We present \DATANAME{}, a benchmark designed to evaluate whether vision-language models can connect past observations, current spatial relationships, and future consequences in challenging driving scenarios. By combining object-centric temporal prompting, six complementary task families, and linked MCQ–OEQ evaluation grounded in structured driving annotations, \DATANAME{} enables fine-grained diagnosis beyond broad scene-description quality. Across 12 proprietary, open-weight, and driving-specialized models, our results show that current systems remain far from reliable: even the strongest model answers fewer than half of the multiple-choice questions correctly, while detailed analysis reveals pronounced gaps between explanation quality and safety-critical judgments, and between recalling past object positions and reconstructing motion history. Together, these findings highlight the need to evaluate driving VLMs through precise spatial and temporal judgments in addition to fluent scene-level explanations. We hope \DATANAME{} will serve as a shared evaluation resource for the research community, supporting systematic comparison of models and diagnosis of their spatiotemporal limitations in challenging driving scenarios.

\subsection*{AI use statement}
In this work, we used generative AI tools during dataset construction solely to verbalize computed labels as natural-language answers. The resulting dataset was subsequently verified by human reviewers. We also used generative AI tools to create small icons for scientific figures; the overall figures were designed and edited by the authors in SVG format. Additionally, we used generative AI tools to polish the writing, and the manuscript underwent multiple rounds of proofreading and revision by the authors. Beyond these uses, we did not use generative AI for any other tasks requiring disclosure. We have reviewed all AI-assisted work. We take responsibility for the final content of this work, including text, claims or artifacts produced with the aid of generative AI.

\subsection*{Ethics statement}
This work evaluates vision-language models using existing driving datasets, with reference answers grounded in dataset annotations and verified by human reviewers. Despite these measures, annotation errors and biases inherited from the source datasets may remain. The benchmark is intended for research evaluation, and its results should be used responsibly, with careful consideration of these limitations and the safety requirements of real-world autonomous driving.

\subsection*{Reproducibility statement}

To support reproducibility, we describe the data sources, preprocessing procedures, and annotation-grounded question and answer construction in Section~\ref{sec:data}. The evaluated models, inference settings, and evaluation metrics are documented in Section~\ref{sec:experiments} and Appendix~\ref{sec:additional_experiments}. Question templates are provided in Appendix~\ref{sec:question_templates}. 



\bibliography{iclr2027_conference}
\bibliographystyle{iclr2027_conference}

\clearpage
\appendix
\appendix
\section*{Appendix}

\startcontents[appendices]
\printcontents[appendices]{}{1}{}

\section{Additional Benchmark Design}

\subsection{WOD-E2E Preprocessing and Label Extraction}
\label{sec:waymo-preprocessing}
For WOD-E2E, several object-centric signals required by our question templates are not directly provided by the released annotations. We therefore construct them through an additional preprocessing pipeline. Specifically, we apply SAM3 with ten road-user prompts to the front-camera images, filter detections by confidence, mask quality, and minimum area, and retain up to ten high-quality objects per frame according to confidence, proximity, and driving relevance. The resulting boxes are deduplicated and associated across consecutive frames using heuristic tracking, from which we derive relative motion, range rate, time-to-collision, relative-position sectors, lane-position bins, and occlusion proxies.

Object distance is not obtained from Waymo 3D labels. Instead, for a detected box $(x_1,y_1,x_2,y_2)$, we take its bottom-center pixel

$$
u=\frac{x_1+x_2}{2}, \qquad v=y_2,
$$

and back-project it through the calibrated front camera. The corresponding camera-frame ray is

$$
\mathbf{r}_c=
\begin{bmatrix}
1\\
-\frac{u-c_x}{f_x}\\
-\frac{v-c_y}{f_y}
\end{bmatrix},
$$

where $(f_x,f_y,c_x,c_y)$ are the camera intrinsic parameters. Using the camera-to-vehicle rotation $\mathbf{R}_{vc}$, we obtain $\mathbf{r}*v=\mathbf{R}*{vc}\mathbf{r}_c$ and intersect this ray with the assumed ground plane $z=0$. If $\mathbf{o}_v$ denotes the camera origin in vehicle coordinates, the intersection scale and point are

$$
\lambda=-\frac{o_z}{r_{v,z}},
\qquad
\mathbf{p}=\mathbf{o}_v+\lambda\mathbf{r}_v,
$$

from which the planar object distance is estimated as

$$
d=\sqrt{p_x^2+p_y^2}.
$$

This is therefore a flat-ground ray-intersection estimate, assuming that the bottom center of the detected box corresponds approximately to the object's contact point with the road surface.

We additionally extract camera calibration and timestamps, ego past position, velocity, and acceleration, ego future trajectory and driving intent, preference trajectories, and released long-tail scenario-cluster labels directly from WOD-E2E (Figure ~\ref{fig:pipeline}(c)). These signals are combined with the derived object-level annotations to support rule-based answer construction for the Waymo split of \DATANAME{}.

\subsection{Question Template Design}
Each instance contains five consecutive 360$^\circ$ views formed by stitching synchronized camera images and sampled at 2Hz, with four context frames followed by a query frame in which a bounding box identifies the target for object-centric questions. To support capability-level analysis, we manually design question templates aligned with the task families introduced in Section~\ref{sec:Task Definition}, each specifying the question wording and expected answer structure. For example, spatial-perception templates cover trajectory direction, relative position to the ego vehicle, distance, lane occupancy, occlusion state, and ground contact. Representative multiple-choice templates include \textit{What is the spatial relationship between \texttt{<obj>} and the ego vehicle?} and \textit{What is the approximate distance between \texttt{<obj>} and the ego vehicle?}, with discretized answer choices derived from the underlying annotations. For each five-frame temporal unit, we select a valid target object and instantiate the corresponding template with the highlighted traffic participant in the query frame. A template is applied only when the required visual evidence and structured annotations are available. The same object may support multiple complementary questions, allowing us to probe different aspects of its spatial and temporal state while maintaining a consistent visual context. Free-response templates combine related capabilities into a more comprehensive query, such as describing the target object's relative position, direction of motion, and approximate distance to the ego vehicle in a single response.

\subsection{Object-centric Temporal Prompting}
As illustrated in Figure ~\ref{fig:pipeline}(d), our prompting scheme is designed to reflect the inherently dynamic nature of autonomous driving. We first construct a unified 360$^\circ$ representation of each driving observation by stitching the synchronized multi-camera views provided by the source datasets. This gives the model a more complete view of the surrounding environment rather than restricting evaluation to a single front-facing camera. While some existing VQA benchmarks evaluate models using a single static image, such a setting is insufficient for many driving-related questions. Unlike some other embodied agent tasks, autonomous driving is fundamentally temporal: properties such as whether a nearby vehicle is moving, accelerating, approaching the ego vehicle, or changing lanes often cannot be determined reliably from an isolated frame. We therefore exploit the short driving clips provided by both source datasets and standardize their sampling rate to 2Hz. Each benchmark instance contains five consecutive observations separated by 0.5s. The first four frames provide short-term temporal context, while the fifth serves as the query frame. For object-centric questions, we visually highlight the target traffic participant with a bounding box in the query frame and ask the model to reason about that object using both its current spatial configuration and the preceding temporal context. Other questions may instead target broader scene-level properties. This formulation constrains the model's attention to a clearly defined subject while preserving the temporal evidence needed for reasoning about motion and interaction. Compared with broad scene-level prompting, it also reduces ambiguity in both the question and its expected answer, making model behavior easier to interpret and evaluate consistently.

\subsection{Trustworthy Ground Truth Generation}
To provide reliable supervision, \DATANAME{} grounds its question templates in original dataset annotations, restricting their scope to questions supported by observable or annotated  (Figure ~\ref{fig:pipeline}(e)). Ground truth is derived hierarchically from geometric signals to driving semantics: annotated longitudinal and lateral offsets between the target object and ego vehicle are converted into distance and bearing, then mapped to labels such as \emph{front}, \emph{front-left}, and \emph{rear-right}. Higher-level labels, such as risk, combine relative position, distance, motion direction, speed, and interaction with the ego trajectory. For example, an object directly ahead, approaching at high relative speed within a collision-relevant region, receives a higher risk label than one moving away. For free-form questions, a language model acts solely as a \emph{verbalizer}, composing the source annotations, derived labels, and corresponding multiple-choice answers into natural-language references. This pipeline supports scalable automated construction while minimizing unsupported or hallucinated supervision and preserving traceability from each answer to the underlying driving data.

\begin{table}[ht]
\centering
\caption{Agreement of different automatic evaluation methods with human judgments.}
\label{tab:judge_alignment}
\begin{tabular}{lccc}
\toprule
\textbf{Method} & \textbf{Pearson} & \textbf{Spearman} & \textbf{Mean $|\Delta|$} \\
\midrule
BLEU & $-0.290$ & $-0.120$ & $7.573$ \\
METEOR & $0.098$ & $0.356$ & $5.580$ \\
BLEURT & $0.193$ & $0.195$ & --- \\
Qwen3-VL-32B (text-only) & $0.780$ & $0.804$ & $1.320$ \\
GPT-5.5 (vision) & $0.870$ & $0.781$ & $0.880$ \\
Qwen3-VL-32B (vision) & $0.788$ & $0.845$ & $0.900$ \\
\bottomrule
\end{tabular}
\end{table}

\subsection{Evaluation with LVLM Judges} 
Traditional machine translation metrics such as BLEU \citep{papineni2002bleu} and METEOR \citep{banerjee2005meteor} primarily measure lexical or surface-level similarity and are therefore poorly suited for evaluating open-ended driving reasoning, where semantically equivalent answers may differ substantially in wording \citep{nie2024reason2drive}. We instead employ a scalable model-based judge for free-response evaluation, providing the judge with the visual observations, question, reference answer, and model response so that correctness can be assessed in the context of the underlying driving scene.

We note that the choice of judge itself requires validation. For example, \citet{chen2025automated} report that a text-only LLM judge achieved stronger human alignment than the vision-language judge evaluated in CODA-LM. Rather than assuming that multimodal judging is inherently superior, we compare conventional NLP metrics, text-only LLM judges, and LVLM judges against human annotations on a held-out subset of \DATANAME{}. As shown in Table~\ref{tab:judge_alignment}, contemporary LVLM judges achieve strong agreement with human evaluation while retaining direct access to the visual evidence required by our spatial and temporal questions. We evaluate both proprietary and open-weight judges and adopt \texttt{Qwen3-VL-32B} as our default judge due to its competitive human alignment and greater reproducibility and accessibility.

\subsection{More on Quality Control} 
\label{sec:quality_control}

\subsubsection{Multiple-choice questions}
Each multiple-choice template in the pre-review set of \DATANAME was represented by a sample of 50 items.
Across all 31 MCQ templates, this review set contained 1{,}550 questions.
A template was retained only when the sample had a well-posed question, a usable option set, and ground truth that could be supported from the frames.
Sixteen templates passed and fifteen failed, a pass rate of $16/31 = 51.6\%$.
Because every template contributes the same number of items, the same rate applies to the questions themselves: $800/1{,}550$ ($51.6\%$) were kept and $750/1{,}550$ ($48.4\%$) were removed.
A failed template was removed in full, rather than edited item by item.
Table~\ref{tab:mcq-qc} groups the removed templates by the reason recorded in the review.

\begin{table}[h]
\centering
\caption{Multiple-choice templates removed after manual review of the 50-item sample. Each template contains 50 questions.}
\label{tab:mcq-qc}
\begin{tabular}{llp{0.40\linewidth}}
\toprule
Template & Question & Reason for removal \\
\midrule
SC-1 & Weather and environmental condition & No suitable choice for dark scenes. \\
SC-2 & Time of day & Samples are mostly daytime, so the label is not reliable. \\
SC-3 & Road type and driving environment & Parked cars do not imply a parking lot, and the sample contains no highway. \\
TE-3 & Future lane of the target object & The frames do not provide enough lane information. \\
TE-4 & Ego response to the target object & Ground truth is incorrect. \\
TE-5 & Next traffic event & Ground truth is incorrect and the question requires lane identification. \\
SC-4 & Traffic density & \\
SC-5 & Baseline scene risk & \\
SP-4 & Lane of the target object & \\
SP-6 & Whether the target is on the drivable plane &  \\
SU-3 & Risk posed by the target object & Rejected in manual review for insufficient/unsatisfactory evidence.\\
SU-5 & Geometrically feasible maneuver & \\
SU-6 & Spatial density around the ego vehicle & \\
TE-2 & Time to collision under constant motion & \\
TM-1 & Objects that are no longer visible & \\
\bottomrule
\end{tabular}
\end{table}

\subsubsection{Open-ended questions}
The same template review removed one open-ended template, TM-6 (50 questions), because its descriptions of recent motion were inaccurate.
Five templates remained: SC-6, SP-7, SU-7, TE-6, and TRJ-7, for 250 questions in total.
Their original labels had been derived from rules or from multiple-choice annotations.
nuScenes does not provide weather or lane labels, and comparison with the frames showed that these labels were often wrong.

We then rewrote the labels in two stages.
First, we drew a stratified sample of 50 questions, 10 from each template ($20\%$ of the open-ended set), and scored each GPT-5.5 answer by hand on a 0--10 scale against the images (mean $7.74$).
The score determined the rewrite.
On $19/50$ items ($38\%$) the human score was 10, and the model answer replaced the ground truth unchanged.
On $17/50$ ($34\%$) the score was 7--9: a vision-language model (Qwen3-VL-32B-Instruct) corrected the model answer using the human note and the frames, and that correction became the new ground truth.
On the remaining $14/50$ ($28\%$) the score was below 7, so the same model revised the original ground truth from the note and the frames, without starting from the model answer.

Second, the other 200 questions were audited by the same vision-language model.
It marked $189/200$ ($94.5\%$) inconsistent with the images and proposed a replacement, which we accepted, and marked $11/200$ ($5.5\%$) acceptable.
Altogether, $239/250$ open-ended labels ($95.6\%$) were rewritten.
The most frequent audit errors were wrong object identity, unsafe driving claims, incorrect distance, incorrect risk, and wrong weather or lighting.

\section{More Dataset Statistics}
Here we provide a table showing unique scene counts of \DATANAME. 
\begin{table}[ht]
\centering
\caption{nuScenes question and scene counts by task family. Scene counts use \texttt{scene\_id}. Scenes are reused across question IDs, so family scene counts do not sum to the split total (310 scenes appear in more than one family).}
\label{tab:nuscenes-v6-counts}
\begin{tabular}{lrrrrl}
\toprule
Task family & Questions & MCQ & OEQ & Scenes & Question IDs \\
\midrule
Scene context            &   50 &   0 &  50 &  50 & SC-6 \\
Spatial perception       &  250 & 200 &  50 & 242 & SP-1, SP-2, SP-3, SP-5, SP-7 \\
Spatial understanding    &  200 & 150 &  50 & 186 & SU-1, SU-2, SU-4, SU-7 \\
Temporal memory          &  200 & 200 &   0 & 187 & TM-2, TM-3, TM-4, TM-5 \\
Temporal extrapolation   &  100 &  50 &  50 & 100 & TE-1, TE-6 \\
Trajectory               &  350 & 200 & 150 & 316 & TRJ-1--TRJ-7 \\
\midrule
Split total              & 1150 & 800 & 350 & 761 & 23 IDs \\
\bottomrule
\end{tabular}
\end{table}

\begin{table}[ht]
\centering
\caption{Waymo question and scene counts by task family. Scene counts use \texttt{scene\_id}. The split also contains 691 unique bundles (\texttt{scene\_id} plus \texttt{group\_id}). Family scene counts do not sum to the split total (183 scenes appear in more than one family).}
\label{tab:waymo-v1-counts}
\begin{tabular}{lrrrrl}
\toprule
Task family & Questions & MCQ & OEQ & Scenes & Question IDs \\
\midrule
Scene context            &  200 & 150 &  50 & 117 & SC-1, SC-2, SC-3, SC-4 \\
Spatial perception       &  150 & 100 &  50 & 108 & SP-2, SP-3, SP-3a \\
Temporal memory          &  150 & 150 &   0 & 109 & TM-2, TM-3, TM-5 \\
Temporal extrapolation   &  200 & 200 &   0 & 158 & TE-1, TE-2, TE-4, TE-5 \\
Trajectory               &  500 & 300 & 200 & 135 & TRJ-1--TRJ-10 \\
\midrule
Split total              & 1200 & 900 & 300 & 335 & 24 IDs \\
\bottomrule
\end{tabular}
\end{table}

\section{Additional Experimental Details}
\label{sec:additional_experiments}

This appendix records the evaluated models, decoding settings, and model-specific exceptions used in the main experiments. The shared question template is described separately in Appendix~\ref{sec:question_templates}.

\subsection{Evaluated Models}
\label{sec:app_models}

Table~\ref{tab:app_models} lists every model reported in the main results. Proprietary models are accessed through official APIs (OpenAI Chat Completions for GPT-5.5; Vertex AI for Gemini and Claude). Open-weight models are loaded from Hugging Face and served with vLLM~\citep{kwon2023efficient} on NVIDIA A6000 GPUs, except Dolphins (official OpenFlamingo inference) and UniAD (end-to-end planner, not a VLM).

Unless noted, we use deterministic or near-deterministic decoding. For models with an explicit reasoning/thinking control, we use a \emph{low} setting so that hidden chain-of-thought does not dominate latency or empty the visible-token budget, while still allowing the model to think. Claude is the exception: \texttt{effort=low} produced little or no reasoning in our setup, so we use \texttt{medium}.

\begin{table}[t]
\centering
\footnotesize
\setlength{\tabcolsep}{4pt}
\caption{Evaluated models. Reasoning knobs are API- or family-specific; ``--'' means none. UniAD is scored only on TRJ-5.}
\label{tab:app_models}
\begin{tabular}{@{}lll@{}}
\toprule
Model & Identifier & Reasoning / thinking \\
\midrule
\multicolumn{3}{@{}l}{\textit{Proprietary LVLMs}} \\
GPT-5.5 & \texttt{gpt-5.5} & \texttt{reasoning\_effort=low} \\
GPT-5.6 & \texttt{gpt-5.6} & \texttt{reasoning\_effort=low} \\
GPT-6-Astra & \texttt{gpt-6-Astra} & \texttt{reasoning\_effort=low} \\
Gemini-3.6-Flash & \texttt{gemini-3.6-flash} & \texttt{thinking\_level=low} \\
Gemini-3.1-Pro & \texttt{gemini-3.1-pro-preview} & \texttt{thinking\_level=low} \\
Claude-Sonnet-5 & \texttt{claude-sonnet-5} & \texttt{effort=medium} \\
\midrule
\multicolumn{3}{@{}l}{\textit{Open-weight VLMs}} \\
Qwen3.6-35B-A3B & \texttt{Qwen3.6-35B-A3B} & thinking off \\
Qwen3-VL-30B-A3B-Thinking & \texttt{Qwen3-VL-30B-A3B-Thinking} & native; after \texttt{</think>} \\
Qwen3-VL-8B-Thinking & \texttt{Qwen3-VL-8B-Thinking} & native; after \texttt{</think>} \\
Qwen3.6-27B & \texttt{Qwen3.6-27B} & thinking off \\
\midrule
\multicolumn{3}{@{}l}{\textit{Driving expert models}} \\
Dolphins & \texttt{gray311/Dolphins} & -- \\
UniAD~2.0 & \texttt{UniAD v2.0 (base e2e)} & n/a (TRJ-5 only) \\
\bottomrule
\end{tabular}
\end{table}

\subsection{Shared Evaluation Protocol}
\label{sec:app_shared_protocol}

All VLMs receive the same visual inputs and the same task instruction whenever the interface allows. Non-SC-6 items use five consecutive 360$^\circ$ multi-camera grid frames; the fifth (query) frame highlights the target with a red box when an object is specified. SC-6 uses up to 40 chronological scene grids. Grid images are resized to at most $640{\times}640$ pixels, except SC-6 frames, which are resized to at most $360{\times}360$ to keep the long clip in context.

\subsection{Model-Specific Settings}
\label{sec:app_special_params}

\paragraph{Proprietary APIs}
GPT family is queried with \texttt{image\_detail=low} and \texttt{max\_completion\_tokens}${}=4096$ for both MCQ and OEQ. The large completion budget is required because GPT-5.x spends hidden tokens on reasoning; a short MCQ cap often returns an empty visible answer. Gemini and Claude images are JPEG-encoded at quality $85$ and resized so the long side is at most $768$ pixels, matching the approximate token budget of OpenAI \texttt{image\_detail=low}. Gemini uses \texttt{thinking\_level=low}; Claude uses \texttt{effort=medium} via Vertex \texttt{output\_config}. Proprietary SC-6 inputs use all $40$ frames.

\paragraph{Open-weight Qwen models}
Qwen3.6-27B and Qwen3.6-35B-A3B are run with chat-template thinking \emph{off} (\texttt{enable\_thinking=false}) so they remain non-reasoning open-weight baselines rather than a second thinking family. For these models and Qwen2.5-VL-3B, SC-6 is evenly subsampled to $8$ frames to fit the $8$k context window.

\paragraph{Thinking variants}
Qwen3-VL Thinking models keep native chain-of-thought. We raise both MCQ and OEQ budgets to $4096$ tokens and score only the span after the last \texttt{</think>} tag. The same strip is applied before OEQ judging and trajectory parsing.

\paragraph{Dolphins}
We follow the official inference API: frames are stacked as a video tensor and wrapped as \texttt{USER: <image> is a driving video. \{instruction\} GPT:<answer>}. Decoding uses beam search with $3$ beams, \texttt{do\_sample=false}, \texttt{no\_repeat\_ngram\_size=3}, and at most $64$ / $512$ new tokens for MCQ / OEQ. SC-6 is capped at $16$ frames (the official demo length).

\paragraph{UniAD}
UniAD is not a VLM and is evaluated only on TRJ-5. We take the official base end-to-end planner, convert lidar-frame waypoints to ego coordinates (flip $x$ as in UniAD's planning metric, then apply \texttt{lidar2ego}), and keep the first $5$ of the $6$ predicted $0.5$s steps to match the miniset ground truth.

\subsection{Scoring Notes}
\label{sec:app_scoring}
The reported MCQ accuracy is a micro-average over items. Let $\mathcal{Q}$ be the set of multiple-choice items in a split, and let $c_i \in \{0,1\}$ indicate whether item $i$ is answered correctly. Then
\begin{equation}
\mathrm{Acc} = \frac{1}{|\mathcal{Q}|}\sum_{i \in \mathcal{Q}} c_i
= \frac{\#\{\text{correct MCQ items}\}}{\#\{\text{MCQ items}\}}.
\label{eq:mcq-acc}
\end{equation}
The denominator is $800$ on nuScenes v6 and $900$ on Waymo v1. Free-response items are excluded, including all of nuScenes scene context (SC-6 is OEQ only).

Every MCQ question ID contains $50$ items, so \eqref{eq:mcq-acc} equals the unweighted mean of the per-question-ID accuracies. It is not the unweighted mean of the per-family accuracies: families contain different numbers of MCQ items, and a family with more items contributes more to $\mathrm{Acc}$. Family-level accuracy uses the same pooling, restricted to that family's MCQ items.  For GPT-5.5, the pooled rate is 0.3925 on nuScenes (314/800) and 0.3867 on Waymo (348/900). Averaging the five family accuracies equally would give 0.375 and 0.378.

An MCQ response that cannot be parsed to an option letter is scored incorrect ($c_i = 0$) and remains in the denominator. The random-guess baseline reported alongside accuracy is the mean of $1/|\mathrm{choices}_i|$ over the same items ($0.197991$ on nuScenes v6 and $0.203241$ on Waymo v1). It is not subtracted inside $\mathrm{Acc}$.

Open-ended OEQs (SC-6, SP-7, SU-7, TE-6, TM-6, TRJ-7) are scored with a vision LLM judge (\texttt{Qwen/Qwen3-VL-32B-Instruct}, temperature $0$); the judge sees the same frames as the candidate, with SC-6 subsampled to $8$ frames. Waypoint OEQs (TRJ-5/TRJ-6) are parsed as JSON $[x,y]$ lists and scored with ADE-style L2 metrics. Unparseable MCQ or trajectory outputs are counted as incorrect / failed, not dropped.

\section{Additional Per-task Performance Tables}
The results of fine-grained scores across models in each specific task are in Table~\ref{tab:nusc_subtask_results} and Table~\ref{tab:waymo_subtask_results}.

\begin{table*}[ht]
\centering
\caption{
Fine-grained results on the nuScenes portion of \DATANAME{}.
MCQ accuracy is reported in percentage.
OEQ responses are evaluated by the LVLM judge.
For ADE, lower is better; for all other metrics, higher is better.
Best and second-best results in each row are shown in \textbf{bold} and
\underline{underline}, respectively.
}
\label{tab:nusc_subtask_results}

\scriptsize
\setlength{\tabcolsep}{3pt}

\resizebox{\textwidth}{!}{
\begin{tabular}{lllcccccccccccc}
\toprule
& & &
\multicolumn{6}{c}{\textbf{Proprietary}} &
\multicolumn{5}{c}{\textbf{Open-weight}} &
\multicolumn{1}{c}{\textbf{Driving Expert}} \\
\cmidrule(lr){4-9}
\cmidrule(lr){10-14}
\cmidrule(lr){15-15}

\textbf{ID} &
\textbf{Task} &
\textbf{Metric} &
\textbf{GPT-5.5} &
\textbf{GPT-5.6} &
\textbf{GPT-6} &
\textbf{G-Fl.} &
\textbf{G-Pro} &
\textbf{Claude} &
\textbf{Q35} &
\textbf{Q2.5} &
\textbf{Q30-T} &
\textbf{Q8-T} &
\textbf{Q27} &
\textbf{Dolphins} \\
\midrule

\multicolumn{15}{l}{\textit{Scene Context}} \\
SC-6 & Environment description & Judge $\uparrow$
& \textbf{9.22} & 8.94 & 8.94 & 8.50 & 7.22 & 8.94
& 8.84 & 7.16 & \underline{9.14} & 9.10 & 8.94 & 4.58 \\

\addlinespace
\multicolumn{15}{l}{\textit{Space Perception}} \\
SP-1 & Trajectory direction & Acc. (\%) $\uparrow$
& 24.0 & 32.0 & \textbf{38.0} & \underline{36.0} & 24.0 & 16.0
& 14.0 & 20.0 & 20.0 & 18.0 & 6.0 & 26.0 \\

SP-2 & Relative position & Acc. (\%) $\uparrow$
& 56.0 & 42.0 & \underline{60.0} & \textbf{62.0} & 54.0 & 40.0
& 46.0 & 32.0 & 40.0 & 34.0 & 52.0 & 10.0 \\

SP-3 & Object distance & Acc. (\%) $\uparrow$
& 40.0 & 42.0 & 44.0 & \underline{50.0} & \textbf{64.0} & 42.0
& 40.0 & 18.0 & 32.0 & 22.0 & 22.0 & 20.0 \\

SP-5 & Occlusion state & Acc. (\%) $\uparrow$
& 36.0 & \underline{42.0} & \textbf{44.0} & 40.0 & 40.0 & 30.0
& 28.0 & 22.0 & 24.0 & 24.0 & 24.0 & 12.0 \\

SP-7 & Spatial description & Judge $\uparrow$
& \textbf{3.66} & 1.96 & 2.94 & 3.22 & 2.10 & \underline{3.42}
& 2.74 & 0.78 & 2.16 & 2.28 & 2.86 & 1.12 \\

\addlinespace
\multicolumn{15}{l}{\textit{Space Understanding}} \\
SU-1 & Physical constraints & Acc. (\%) $\uparrow$
& 32.0 & 32.0 & 26.0 & 28.0 & 30.0 & 32.0
& \underline{38.0} & 36.0 & 36.0 & \textbf{42.0} & 36.0 & 26.0 \\

SU-2 & Drivable region & Acc. (\%) $\uparrow$
& 22.0 & \textbf{32.0} & 24.0 & \underline{26.0} & \underline{26.0} & 16.0
& 6.0 & 18.0 & 10.0 & 22.0 & 0.0 & 16.0 \\

SU-4 & Lane-change clearance & Acc. (\%) $\uparrow$
& 22.0 & 30.0 & \textbf{40.0} & 18.0 & 28.0 & 34.0
& 30.0 & \underline{36.0} & 26.0 & 24.0 & 20.0 & 26.0 \\

SU-7 & Overall spatial situation & Judge $\uparrow$
& \underline{7.18} & 6.68 & \textbf{7.44} & 6.36 & 3.90 & 6.48
& 5.86 & 3.92 & 6.70 & 6.86 & 6.84 & 3.48 \\

\addlinespace
\multicolumn{15}{l}{\textit{Time Extrapolation}} \\
TE-1 & Future object motion & Acc. (\%) $\uparrow$
& 32.0 & \underline{34.0} & \underline{34.0} & 28.0 & \textbf{40.0} & 24.0
& 30.0 & 24.0 & 28.0 & 20.0 & 16.0 & 16.0 \\

TE-6 & Future scene evolution & Judge $\uparrow$
& 5.44 & 4.46 & 5.36 & \textbf{6.98} & 4.32 & 5.14
& 4.16 & 1.76 & 4.20 & 4.58 & \underline{5.82} & 1.66 \\

\addlinespace
\multicolumn{15}{l}{\textit{Time Memory}} \\
TM-2 & Previous object position & Acc. (\%) $\uparrow$
& \textbf{72.0} & \textbf{72.0} & \underline{70.0} & 56.0 & 68.0 & 52.0
& 56.0 & 42.0 & 30.0 & 22.0 & 32.0 & 10.0 \\

TM-3 & Recent motion pattern & Acc. (\%) $\uparrow$
& 18.0 & 16.0 & 20.0 & 18.0 & \underline{22.0} & 20.0
& 14.0 & \textbf{30.0} & 20.0 & 18.0 & 2.0 & 14.0 \\

TM-4 & Previously visible objects & Acc. (\%) $\uparrow$
& \underline{32.0} & 28.0 & 26.0 & \underline{32.0} & 22.0 & 26.0
& \textbf{36.0} & 20.0 & 20.0 & 8.0 & 8.0 & 26.0 \\

TM-5 & Previous lane occupancy & Acc. (\%) $\uparrow$
& 42.0 & 32.0 & \textbf{52.0} & \underline{44.0} & 38.0 & 28.0
& 42.0 & 38.0 & 36.0 & 32.0 & 26.0 & 26.0 \\

\addlinespace
\multicolumn{15}{l}{\textit{Trajectory Prediction}} \\
TRJ-1 & Past ego maneuver & Acc. (\%) $\uparrow$
& \underline{42.0} & 40.0 & \textbf{50.0} & \underline{42.0} & 38.0 & 34.0
& 30.0 & 26.0 & 26.0 & 22.0 & 30.0 & 10.0 \\

TRJ-2 & Future ego maneuver & Acc. (\%) $\uparrow$
& \textbf{56.0} & 44.0 & \underline{54.0} & 38.0 & 32.0 & 28.0
& 32.0 & 22.0 & 28.0 & 26.0 & 2.0 & 18.0 \\

TRJ-3 & Future relative position & Acc. (\%) $\uparrow$
& \underline{52.0} & 46.0 & \textbf{90.0} & 46.0 & 38.0 & 26.0
& 30.0 & 28.0 & 30.0 & 24.0 & 2.0 & 18.0 \\

TRJ-4 & Future speed trend & Acc. (\%) $\uparrow$
& 50.0 & 52.0 & \underline{56.0} & \textbf{58.0} & 40.0 & 22.0
& 34.0 & 26.0 & 22.0 & 16.0 & 2.0 & 22.0 \\

TRJ-5 & Future trajectory & ADE $\downarrow$
& \textbf{0.70} & 0.77 & \underline{0.72} & 0.74 & 1.91 & 0.91
& 6.66 & -- & 5.51 & 6.08 & 10.13 & 8.59 \\

TRJ-6 & Conditioned future trajectory & ADE $\downarrow$
& \underline{0.61} & 0.62 & \textbf{0.49} & 0.63 & 0.92 & 0.83
& 8.19 & -- & 8.73 & 9.91 & 1.49 & 7.73 \\

TRJ-7 & Future path description & Judge $\uparrow$
& \underline{7.90} & 6.20 & 7.36 & 7.42 & 5.10 & \textbf{8.12}
& 5.02 & 4.46 & 5.04 & 4.70 & 5.50 & 5.34 \\

\bottomrule
\end{tabular}
}

\vspace{2pt}
\parbox{\textwidth}{\scriptsize
GPT-5.6: GPT-5.6-Sol;
GPT-6: GPT-6-Astra;
G-Fl.: Gemini-3.6-Flash;
G-Pro: Gemini-3.1-Pro;
Q35: Qwen3.6-35B-A3B;
Q2.5: Qwen2.5-VL-3B;
Q30-T: Qwen3-VL-30B-A3B-Thinking;
Q8-T: Qwen3-VL-8B-Thinking;
Q27: Qwen3.6-27B.
}
\end{table*}

\begin{table*}[ht]
\centering
\caption{
Fine-grained results on the Waymo E2E split of \DATANAME{}.
MCQ accuracy is reported in percentage.
OEQ responses are evaluated by the LVLM judge.
For ADE, lower is better; for all other metrics, higher is better.
Best and second-best results in each row are shown in \textbf{bold} and
\underline{underline}, respectively.
}
\label{tab:waymo_subtask_results}

\scriptsize
\setlength{\tabcolsep}{3pt}

\resizebox{\textwidth}{!}{
\begin{tabular}{lllccccccccccccc}
\toprule
& & &
\multicolumn{6}{c}{\textbf{Proprietary}} &
\multicolumn{6}{c}{\textbf{Open-weight}} &
\multicolumn{1}{c}{\textbf{Driving Expert}} \\
\cmidrule(lr){4-9}
\cmidrule(lr){10-15}
\cmidrule(lr){16-16}

\textbf{ID} &
\textbf{Task} &
\textbf{Metric} &
\textbf{GPT-5.5} &
\textbf{GPT-5.6} &
\textbf{GPT-6} &
\textbf{G-Fl.} &
\textbf{G-Pro} &
\textbf{Claude} &
\textbf{Q35} &
\textbf{Q2.5} &
\textbf{Q30-T} &
\textbf{Q8-T} &
\textbf{Q8-I} &
\textbf{Q27} &
\textbf{Dolphins} \\
\midrule

\multicolumn{16}{l}{\textit{Scenario / Corner Case}} \\
SC-1 & Scene description & Acc. (\%) $\uparrow$
& 44.0 & \underline{46.0} & \textbf{48.0} & 40.0 & 38.0 & 40.0
& 32.0 & 42.0 & 36.0 & 38.0 & 44.0 & 40.0 & 32.0 \\

SC-2 & Main risk source & Acc. (\%) $\uparrow$
& 28.0 & 24.0 & \textbf{36.0} & 20.0 & 30.0 & \textbf{36.0}
& 22.0 & 24.0 & 26.0 & \underline{34.0} & 26.0 & 30.0 & 20.0 \\

SC-3 & Preferred ego action & Acc. (\%) $\uparrow$
& \textbf{48.0} & \underline{46.0} & 44.0 & 42.0 & 44.0 & 38.0
& 36.0 & \underline{46.0} & 38.0 & 42.0 & 38.0 & 44.0 & 26.0 \\

SC-4 & Scene/risk/action explanation & Judge $\uparrow$
& \underline{8.08} & 7.80 & 7.76 & 7.46 & 4.22 & \textbf{8.28}
& 7.14 & 4.02 & 6.92 & 6.98 & 7.40 & 7.76 & 2.38 \\

\addlinespace
\multicolumn{16}{l}{\textit{Space Perception}} \\
SP-2 & Relative position & Acc. (\%) $\uparrow$
& 32.0 & 36.0 & 34.0 & \underline{40.0} & 34.0 & 32.0
& \textbf{44.0} & 36.0 & 26.0 & 28.0 & 14.0 & \underline{40.0} & 16.0 \\

SP-3 & Object distance & Acc. (\%) $\uparrow$
& 50.0 & 36.0 & 54.0 & \textbf{62.0} & 48.0 & 52.0
& 46.0 & 18.0 & 38.0 & 40.0 & 22.0 & \underline{56.0} & 20.0 \\

SP-3a & Distance estimation & Judge $\uparrow$
& 5.14 & 5.42 & 5.82 & \underline{5.84} & 4.72 & \textbf{6.28}
& 5.12 & 1.40 & 3.98 & 2.70 & 3.16 & 4.28 & 0.20 \\

\addlinespace
\multicolumn{16}{l}{\textit{Time Extrapolation}} \\
TE-1 & Future object motion & Acc. (\%) $\uparrow$
& 28.0 & 30.0 & 26.0 & 20.0 & \underline{32.0} & 26.0
& \textbf{36.0} & 24.0 & 28.0 & 30.0 & 28.0 & 22.0 & 24.0 \\

TE-2 & Collision timing & Acc. (\%) $\uparrow$
& 36.0 & \textbf{42.0} & \textbf{42.0} & \textbf{42.0}
& \underline{40.0} & 38.0 & 38.0 & \textbf{42.0}
& 38.0 & 18.0 & 38.0 & 38.0 & 20.0 \\

TE-4 & Ego response & Acc. (\%) $\uparrow$
& 28.0 & \textbf{34.0} & 22.0 & 26.0 & 26.0 & \underline{30.0}
& 28.0 & 20.0 & 22.0 & 16.0 & 16.0 & 26.0 & 20.0 \\

TE-5 & Next traffic event & Acc. (\%) $\uparrow$
& 28.0 & 24.0 & 22.0 & \textbf{34.0} & 26.0 & \underline{32.0}
& 18.0 & \textbf{34.0} & 26.0 & 30.0 & 22.0 & 20.0 & 18.0 \\

\addlinespace
\multicolumn{16}{l}{\textit{Time Memory}} \\
TM-2 & Previous object position & Acc. (\%) $\uparrow$
& 30.0 & 26.0 & 30.0 & \underline{42.0} & 32.0 & \underline{42.0}
& 32.0 & \textbf{52.0} & 10.0 & 8.0 & 10.0 & 34.0 & 20.0 \\

TM-3 & Recent motion pattern & Acc. (\%) $\uparrow$
& 24.0 & \underline{30.0} & 22.0 & 22.0 & 18.0 & 20.0
& 20.0 & 26.0 & \textbf{32.0} & 18.0 & 28.0 & 18.0 & 16.0 \\

TM-5 & Previous lane occupancy & Acc. (\%) $\uparrow$
& 40.0 & \underline{42.0} & \textbf{44.0} & 38.0 & 36.0 & 34.0
& 34.0 & 38.0 & 18.0 & 38.0 & 34.0 & 30.0 & 26.0 \\

\addlinespace
\multicolumn{16}{l}{\textit{Trajectory Prediction}} \\
TRJ-1 & Past ego maneuver & Acc. (\%) $\uparrow$
& 62.0 & 54.0 & \underline{72.0} & 60.0 & \underline{72.0} & \textbf{74.0}
& 36.0 & 54.0 & 44.0 & 40.0 & 44.0 & 50.0 & 18.0 \\

TRJ-2 & Future ego maneuver & Acc. (\%) $\uparrow$
& 56.0 & 42.0 & \textbf{64.0} & 54.0 & \underline{60.0} & 54.0
& 46.0 & 52.0 & 54.0 & 52.0 & 54.0 & 24.0 & 18.0 \\

TRJ-3 & Future relative position & Acc. (\%) $\uparrow$
& 62.0 & 60.0 & \textbf{76.0} & 64.0 & \underline{68.0} & 56.0
& 50.0 & 58.0 & 58.0 & 48.0 & 38.0 & 2.0 & 18.0 \\

TRJ-4 & Future speed trend & Acc. (\%) $\uparrow$
& 36.0 & \underline{46.0} & 40.0 & 34.0 & \textbf{48.0} & 30.0
& 30.0 & 28.0 & 26.0 & 28.0 & 28.0 & 2.0 & 22.0 \\

TRJ-5 & Future trajectory & ADE $\downarrow$
& 23.44 & 23.26 & 23.04 & 22.59 & 24.01 & 21.85
& 20.00 & -- & \textbf{19.23} & 38.69 & 33.58 & 29.47 & \underline{19.83} \\

TRJ-6 & Conditioned future trajectory & ADE $\downarrow$
& \textbf{4.25} & 4.36 & \underline{4.25} & 4.47 & 4.44 & 4.46
& 17.10 & -- & 4.44 & 6.04 & 5.69 & 11.28 & 17.02 \\

TRJ-7 & Future path description & Judge $\uparrow$
& \underline{6.22} & \underline{6.22} & \textbf{6.26} & 5.98 & 4.28 & 5.40
& 5.44 & 3.96 & 5.02 & 4.38 & 4.30 & 6.00 & 3.30 \\

TRJ-8 & Best candidate trajectory & Acc. (\%) $\uparrow$
& 34.0 & 30.0 & \underline{38.0} & 26.0 & 34.0 & \textbf{40.0}
& \textbf{40.0} & \underline{38.0} & 30.0 & \underline{38.0}
& 28.0 & 6.0 & \textbf{40.0} \\

TRJ-9 & Trajectory justification & Judge $\uparrow$
& \textbf{8.56} & 6.46 & 7.24 & 5.52 & 2.88 & \underline{8.08}
& 5.60 & 0.90 & 3.98 & 4.80 & 4.98 & 5.42 & 1.54 \\

TRJ-10 & Alternative trajectory consequence & Acc. (\%) $\uparrow$
& 30.0 & 30.0 & 28.0 & \underline{32.0} & 28.0 & \underline{32.0}
& 30.0 & \textbf{42.0} & 22.0 & 30.0 & 30.0 & 30.0 & 26.0 \\

\bottomrule
\end{tabular}
}

\vspace{2pt}
\parbox{\textwidth}{\scriptsize
GPT-5.6: GPT-5.6-Sol;
GPT-6: GPT-6-Astra;
G-Fl.: Gemini-3.6-Flash;
G-Pro: Gemini-3.1-Pro;
Q35: Qwen3.6-35B-A3B;
Q2.5: Qwen2.5-VL-3B;
Q30-T: Qwen3-VL-30B-A3B-Thinking;
Q8-T: Qwen3-VL-8B-Thinking;
Q8-I: Qwen3-VL-8B-Instruct;
Q27: Qwen3.6-27B.
}
\end{table*}

\clearpage

\section{Question Templates}
\label{sec:question_templates}

The benchmark taxonomy contains 39 question templates across six families. Tables~\ref{tab:tmpl_sc}--\ref{tab:tmpl_trj} list every template. Columns \textbf{N} and \textbf{W} mark use in the reported nuScenes and Waymo evaluation splits. A blank cell means the template exists in the full catalog but is not used in that split. Waymo reuses IDs SC-1--4 for a different scenario-corner-case family and rewrites TRJ-5/6 for front-view input; those variants are recorded in Table~\ref{tab:tmpl_waymo_extra}.

\begin{table}[ht]
\centering
\scriptsize
\setlength{\tabcolsep}{3pt}
\caption{Scene-context templates. SC-1--5 were dropped from the reviewed nuScenes split.}
\label{tab:tmpl_sc}
\begin{tabular}{@{}llp{0.58\linewidth}cc@{}}
\toprule
ID & Fmt & Template & N & W \\
\midrule
SC-1 & MCQ & What is the primary weather and environmental condition in the current scene? \newline
{\em A}~Clear/Sunny; {\em B}~Rain/Wet; {\em C}~Overcast; {\em D}~Snow/Ice; {\em E}~Fog/Haze & & \\
SC-2 & MCQ & Based on the lighting conditions and sky visibility, what is the estimated time of day? \newline
{\em A}~Daytime; {\em B}~Nighttime; {\em C}~Dawn/Dusk & & \\
SC-3 & MCQ & Which of the following best describes the overall road typology and driving environment? \newline
{\em A}~Urban intersection; {\em B}~Straight urban/suburban road; {\em C}~Highway; {\em D}~Construction zone; {\em E}~Parking lot/off-road & & \\
SC-4 & MCQ & How spatially dense or crowded is the current traffic environment considering all visible road users? \newline
{\em A}~Sparse; {\em B}~Moderate; {\em C}~Dense & & \\
SC-5 & MCQ & What is the baseline safety risk level of the overall scene, independent of any single specific object? \newline
{\em A}~Low; {\em B}~Moderate; {\em C}~High & & \\
SC-6 & OEQ & Describe the overall environment of the scene, including weather, lighting conditions, road type, and notable surrounding activity. & Y & \\
\bottomrule
\end{tabular}
\end{table}

\begin{table}[t]
\centering
\scriptsize
\setlength{\tabcolsep}{3pt}
\caption{Space-perception templates.}
\label{tab:tmpl_sp}
\begin{tabular}{@{}llp{0.58\linewidth}cc@{}}
\toprule
ID & Fmt & Template & N & W \\
\midrule
SP-1 & MCQ & What is the trajectory direction of \texttt{<obj>} in relation to the ego vehicle? \newline
{\em A}~Toward ego; {\em B}~Away from ego; {\em C}~Left-to-right; {\em D}~Right-to-left; {\em E}~Static; {\em F}~Same direction as ego & Y & \\
SP-2 & MCQ & What is the spatial relationship between \texttt{<obj>} and ego vehicle? \newline
{\em A}~In front; {\em B}~Front-left; {\em C}~Front-right; {\em D}~Directly left; {\em E}~Directly right; {\em F}~Rear-left; {\em G}~Rear-right; {\em H}~Behind & Y & Y \\
SP-3 & MCQ & What is the approximate distance between \texttt{<obj>} and ego vehicle? \newline
{\em A}~0--5m; {\em B}~5--15m; {\em C}~15--30m; {\em D}~30--50m; {\em E}~{$>$}50m & Y & Y \\
SP-4 & MCQ & Use the ego vehicle's lanes as reference, which lane is \texttt{<obj>} located in? \newline
{\em A}~Same lane; {\em B}~Left adjacent; {\em C}~Right adjacent; {\em D}~Two lanes left; {\em E}~Two lanes right & & \\
SP-5 & MCQ & What is the occlusion state of \texttt{<obj>}? \newline
{\em A}~Fully visible; {\em B}~Partially occluded ({$<$}50\%); {\em C}~Mostly occluded ($\geq$50\%); {\em D}~Fully occluded but inferred & Y & \\
SP-6 & MCQ & Is \texttt{<obj>} on the drivable ground plane? \newline
{\em A}~On road; {\em B}~On sidewalk; {\em C}~Above the ground & & \\
SP-7 & OEQ & Describe the spatial relationship between the ego vehicle and \texttt{<obj>}, including its relative position, orientation of motion, and approximate distance. & Y & \\
\bottomrule
\end{tabular}
\end{table}

\begin{table}[t]
\centering
\scriptsize
\setlength{\tabcolsep}{3pt}
\caption{Space-understanding templates. This family is nuScenes-only.}
\label{tab:tmpl_su}
\begin{tabular}{@{}llp{0.58\linewidth}cc@{}}
\toprule
ID & Fmt & Template & N & W \\
\midrule
SU-1 & MCQ & What physical constraints affect the ego vehicle given \texttt{<obj>}? \newline
{\em A}~Blocking ego lane; {\em B}~Partially obstructing; {\em C}~No constraint; {\em D}~Dynamic obstruction in ego lane & Y & \\
SU-2 & MCQ & Which region around the ego vehicle is safely drivable? \newline
{\em A}~Front only; {\em B}~Front-left; {\em C}~Front-right; {\em D}~Left only; {\em E}~Right only; {\em F}~None; {\em G}~All directions & Y & \\
SU-3 & MCQ & What is the risk level posed by \texttt{<obj>} to the ego vehicle? \newline
{\em A}~Low; {\em B}~Moderate; {\em C}~High; {\em D}~Critical/imminent collision & & \\
SU-4 & MCQ & Does the ego vehicle have sufficient clearance to change lanes? \newline
{\em A}~Safe left; {\em B}~Safe right; {\em C}~Unsafe both; {\em D}~Safe both & Y & \\
SU-5 & MCQ & Which maneuver is geometrically feasible at this moment? \newline
{\em A}~Maintain lane; {\em B}~Lane change left; {\em C}~Lane change right; {\em D}~Turn left; {\em E}~Turn right; {\em F}~Stop only & & \\
SU-6 & MCQ & How spatially dense is the environment around ego considering the number of objects and their distance to the ego? \newline
{\em A}~Sparse; {\em B}~Moderate; {\em C}~Dense; {\em D}~Highly congested & & \\
SU-7 & OEQ & Describe the overall spatial situation of the ego vehicle in the scene, including existing physical constraints, available drivable space, risk level, and feasible maneuvers. & Y & \\
\bottomrule
\end{tabular}
\end{table}

\begin{table}[t]
\centering
\scriptsize
\setlength{\tabcolsep}{3pt}
\caption{Time-extrapolation templates.}
\label{tab:tmpl_te}
\begin{tabular}{@{}llp{0.58\linewidth}cc@{}}
\toprule
ID & Fmt & Template & N & W \\
\midrule
TE-1 & MCQ & What is the most likely future motion of \texttt{<obj>} over the next few seconds? \newline
{\em A}~Continue straight; {\em B}~Turn left; {\em C}~Turn right; {\em D}~Stop; {\em E}~Cross ego path & Y & Y \\
TE-2 & MCQ & If both vehicles maintain their current motion, when would a collision occur? \newline
{\em A}~No collision; {\em B}~{$>$}10s; {\em C}~5--10s; {\em D}~2--5s; {\em E}~{$<$}2s & & Y \\
TE-3 & MCQ & Which lane will \texttt{<obj>} most likely occupy in the next few seconds? \newline
{\em A}~Remain; {\em B}~Left lane; {\em C}~Right lane; {\em D}~Enter ego lane; {\em E}~Exit roadway & & \\
TE-4 & MCQ & What should be the ego vehicle's primary response to \texttt{<obj>}? \newline
{\em A}~Maintain speed; {\em B}~Slow down; {\em C}~Prepare to stop; {\em D}~Change lane; {\em E}~Yield; {\em F}~Emergency braking & & Y \\
TE-5 & MCQ & What traffic event is most likely to occur next? \newline
{\em A}~Lane change; {\em B}~Vehicle stopping; {\em C}~Pedestrian crossing; {\em D}~Vehicle merging; {\em E}~No significant event & & Y \\
TE-6 & OEQ & Describe how the current scene is likely to evolve over the next few seconds and what actions the ego vehicle should take to maintain safety. & Y & \\
\bottomrule
\end{tabular}
\end{table}

\begin{table}[t]
\centering
\scriptsize
\setlength{\tabcolsep}{3pt}
\caption{Time-memory templates.}
\label{tab:tmpl_tm}
\begin{tabular}{@{}llp{0.58\linewidth}cc@{}}
\toprule
ID & Fmt & Template & N & W \\
\midrule
TM-1 & MCQ & Which objects previously appeared in the scene but are no longer visible in current frame? \newline
{\em A}~None; {\em B}~Pedestrian; {\em C}~Vehicle; {\em D}~Bicycle; {\em E}~Static obstacle; {\em F}~Multiple objects & & \\
TM-2 & MCQ & Where was \texttt{<obj>} located relative to the ego vehicle in the previous frame(s)? \newline
{\em A}~In front; {\em B}~Front-left; {\em C}~Front-right; {\em D}~Left; {\em E}~Right; {\em F}~Behind & Y & Y \\
TM-3 & MCQ & What was the recent motion pattern of \texttt{<obj>}? \newline
{\em A}~Moving straight; {\em B}~Slowing down; {\em C}~Accelerating; {\em D}~Turning left; {\em E}~Turning right; {\em F}~Stopped previously & Y & Y \\
TM-4 & MCQ & How many objects are currently occluded but were previously visible in the previous frame? \newline
{\em A}~None; {\em B}~1; {\em C}~2--3; {\em D}~More than 3 & Y & \\
TM-5 & MCQ & In which lane was \texttt{<obj>} located earlier in the sequence? \newline
{\em A}~Same lane; {\em B}~Left adjacent; {\em C}~Right adjacent; {\em D}~Outside roadway & Y & Y \\
TM-6 & OEQ & Describe the recent history of the scene, including objects that appeared or disappeared, changes in their motion, and any interactions with the ego vehicle. & & \\
\bottomrule
\end{tabular}
\end{table}

\begin{table}[t]
\centering
\scriptsize
\setlength{\tabcolsep}{3pt}
\caption{Trajectory-prediction templates. TRJ-5/6 on Waymo keep the ID but change the visual preamble (front-view, $\sim$2.0s); see Table~\ref{tab:tmpl_waymo_extra}.}
\label{tab:tmpl_trj}
\begin{tabular}{@{}llp{0.58\linewidth}cc@{}}
\toprule
ID & Fmt & Template & N & W \\
\midrule
TRJ-1 & MCQ & Based on the past 5 consecutive 360$^\circ$ multi-camera frames, what best describes the ego vehicle's recent maneuver over those past frames? \newline
{\em A}~Go straight; {\em B}~Slightly veer left; {\em C}~Slightly veer right; {\em D}~Turn left; {\em E}~Turn right; {\em F}~Slow to near stop & Y & Y \\
TRJ-2 & MCQ & Based on the past 5 consecutive 360$^\circ$ multi-camera frames, what is the most likely ego-vehicle maneuver over the next 5 future frames (approximately 2.5 seconds)? \newline
{\em A}--{\em F} as TRJ-1 & Y & Y \\
TRJ-3 & MCQ & Based on the past 5 consecutive 360$^\circ$ multi-camera frames, at the 5th future frame (approximately 2.5 seconds ahead), which relative endpoint region best matches the ego vehicle trajectory in the current ego-centric coordinate system? \newline
{\em A}~Front-left; {\em B}~Front-center; {\em C}~Front-right; {\em D}~Left-side; {\em E}~Right-side; {\em F}~Near current position & Y & Y \\
TRJ-4 & MCQ & Based on the past 5 consecutive 360$^\circ$ multi-camera frames, how does the ego vehicle speed trend evolve over the next 5 future frames (approximately 2.5 seconds)? \newline
{\em A}~Accelerating; {\em B}~Constant speed; {\em C}~Decelerating; {\em D}~Stopping/nearly stopped & Y & Y \\
TRJ-5 & OEQ & Using past ego positions \texttt{<trajectory\_history>} in the ego frame (current ${=}[0,0]$), predict the next 5 future positions as a JSON array of 5 $[x,y]$ points. & Y & $^\dagger$ \\
TRJ-6 & OEQ & Same setup as TRJ-5, with known first future point \texttt{<first\_future\_position>}. Predict only the remaining 4 points. & Y & $^\dagger$ \\
TRJ-7 & OEQ & Based on the past 5 consecutive 360$^\circ$ multi-camera frames, briefly describe the ego vehicle's likely future path over the next 5 future frames (approximately 2.5 seconds). Mention the overall maneuver, endpoint region, and speed trend in 1--2 sentences. & Y & Y \\
\bottomrule
\end{tabular}
\end{table}

\begin{table}[t]
\centering
\scriptsize
\setlength{\tabcolsep}{3pt}
\caption{Templates that appear only on Waymo, or reuse a nuScenes ID with different wording. SC-1--4 here are \emph{scenario-corner-case}, not scene-context. $^\dagger$TRJ-5/6 drop the 360$^\circ$ preamble and the pasted history, and ask for a front-view 2.0s horizon.}
\label{tab:tmpl_waymo_extra}
\begin{tabular}{@{}llp{0.72\linewidth}@{}}
\toprule
ID & Fmt & Waymo template \\
\midrule
SC-1 & MCQ & Which option best describes the current scene? \newline
{\em A}~highway; {\em B}~traffic intersection; {\em C}~construction zone; {\em D}~right turn \\
SC-2 & MCQ & What is the main risk source in this scene? \newline
{\em A}~Road debris; {\em B}~Cut-in vehicle; {\em C}~Crossing pedestrian/cyclist; {\em D}~Unclear lane boundary; {\em E}~No clear immediate risk \\
SC-3 & MCQ & What is the preferred action for the ego vehicle? \newline
{\em A}~Slow down; {\em B}~Keep lane carefully; {\em C}~Change lane; {\em D}~Stop and wait; {\em E}~Speed up \\
SC-4 & OEQ & Briefly describe the scene, the main risk, and the preferred action. \\
SP-3a & OEQ & What is the approximate distance in meters between \texttt{<obj>} and the ego vehicle? \\
TRJ-5$^\dagger$ & OEQ & Based on the past 5 observed front-view frames sampled over approximately 2.0 seconds, predict the next 5 future ego positions as a JSON array of 5 $[x,y]$ points. Current frame is $[0,0]$. \\
TRJ-6$^\dagger$ & OEQ & Same front-view setup, with the first future point given. Predict only the remaining 4 points. \\
TRJ-8 & MCQ & Given the current scene and three candidate ego future trajectories, which trajectory is the best driving choice? \newline
{\em A}--{\em C}~Candidate trajectory A/B/C \\
TRJ-9 & OEQ & Explain why the selected trajectory is the best driving choice in this scene. \\
TRJ-10 & MCQ & What is the most likely consequence if the ego vehicle follows the alternative trajectory to move or turn left, instead of the best trajectory? \newline
{\em A}~Crash into another vehicle; {\em B}~Crash into a VRU; {\em C}~Enter opposing traffic; {\em D}~Drift too far laterally \\
\bottomrule
\end{tabular}
\end{table}



\end{document}